\documentclass{article}

\PassOptionsToPackage{numbers, compress}{natbib}

\usepackage[final]{neurips_2024}

\usepackage[utf8]{inputenc} 
\usepackage[T1]{fontenc}    
\usepackage{url}            
\usepackage{booktabs}       
\usepackage{enumitem}
\usepackage{natbib}
\usepackage{amsfonts}       
\usepackage{nicefrac}       
\usepackage{microtype}      
\usepackage{xcolor}         
\usepackage{enumitem}
\usepackage{xspace}
\usepackage{mathtools}
\usepackage{todonotes}
\usepackage{amssymb,fge}
\usepackage{thm-restate}
\usepackage{wrapfig}
\usepackage{bbm}
\usepackage{mathrsfs}
\usepackage{soul}
\usepackage{array}
\usepackage{multirow}
\usepackage{algorithm}
\usepackage[commentColor=black,beginLComment=/*~, endLComment=~*/]{algpseudocodex}
\usepackage{subcaption}
\usepackage{pifont}
\usepackage{tikz-3dplot}
\usepackage{pgfplots}
\pgfplotsset{compat=newest}
\usepackage{tikzscale}
\usepackage{comment} 

\usepackage[final,pagebackref=true]{hyperref} 
\usepackage{cleveref}
\usepackage{wasysym}
\usepackage{acronym}
\usepackage{xfrac}
\acrodef{FBD}[Fr\'echet Biological Distance]{FBD}

\usepackage{amsthm}
\usepackage{amsmath,amsfonts,bm}
\makeatletter
\newtheorem*{rep@theorem}{\rep@title}
\newcommand{\newreptheorem}[2]{%
\newenvironment{rep#1}[1]{%
 \def\rep@title{#2 \ref{##1}}%
 \begin{rep@theorem}}%
 {\end{rep@theorem}}}
\makeatother

\newcommand{\Nstar}{\mathbb{N}^\star}
\newcommand{\sphere}{\mathbb{S}}
\newcommand{\posOrthant}{\sphere_{+}}
\newcommand{\tangent}{\mathcal{T}}
\newcommand{\diff}{\mathrm{d}}
\DeclarePairedDelimiter{\norm}{\lVert}{\rVert}

\newreptheorem{theorem}{Theorem}

\newenvironment{proofsketch}{%
  \proof}{\endproof}

\definecolor{myred}{RGB}{215,48,39}
\definecolor{mygreen}{RGB}{26,152,80}

\newcommand{\halfmark}{\textcolor{gray}{\checkmark\kern-1.1ex\raisebox{.7ex}{\rotatebox[origin=c]{125}{--}}}}
\usepackage[framemethod=TikZ]{mdframed}
\mdfdefinestyle{MyFrame}{%
    linecolor=black,
    outerlinewidth=0.5pt,
    roundcorner=1pt,
    innertopmargin=2pt, 
    innerbottommargin=2pt, 
    innerrightmargin=7pt,
    innerleftmargin=7pt,
    backgroundcolor=black!0!white}

\mdfdefinestyle{MyFrame2}{%
    linecolor=white,
    outerlinewidth=1pt,
    roundcorner=2pt,
    innertopmargin=0.5pt,
    innerbottommargin=0.5pt,
    innerrightmargin=10pt,
    innerleftmargin=10pt,
    backgroundcolor=black!3!white}

\mdfdefinestyle{MyFrameEq}{%
    linecolor=white,
    outerlinewidth=0pt,
    roundcorner=0pt,
    innertopmargin=0pt,
    innerbottommargin=0pt,
    innerrightmargin=7pt,
    innerleftmargin=7pt,
    backgroundcolor=black!3!white}

\newcommand{\RNum}[1]{\uppercase\expandafter{\romannumeral #1\relax}}

\newcommand{\R}{\mathcal{R}}

\newcommand{\vertiii}[1]{{\left\vert\kern-0.25ex\left\vert\kern-0.25ex\left\vert #1 
    \right\vert\kern-0.25ex\right\vert\kern-0.25ex\right\vert}}
\newcommand{\vertiiii}[1]{{\vert\kern-0.25ex\vert\kern-0.25ex\vert #1 
    \vert\kern-0.25ex\vert\kern-0.25ex\vert}}

\usepackage{mathtools}

\DeclareMathOperator*{\argmin}{\arg\!\min}
\DeclareMathOperator*{\argmax}{\arg\!\max}
\newcommand{\xhdr}[1]{{\noindent\bfseries #1}.}
\newcommand{\cut}[1]{}
\newcommand{\CITE}{\textcolor{red}{CITE}}

\newcommand{\removelatexerror}{\let\@latex@error\@gobble}

\def\eqref#1{Eq.~\ref{#1}}

\def\1{\bm{1}}

\DeclareMathAlphabet{\mathsfit}{\encodingdefault}{\sfdefault}{m}{sl}
\SetMathAlphabet{\mathsfit}{bold}{\encodingdefault}{\sfdefault}{bx}{n}

\def\gA{{\mathcal{A}}}
\def\gB{{\mathcal{B}}}

\def\gD{{\mathcal{D}}}

\def\gL{{\mathcal{L}}}
\def\gM{{\mathcal{M}}}
\def\gN{{\mathcal{N}}}

\def\gP{{\mathcal{P}}}

\def\gT{{\mathcal{T}}}
\def\gU{{\mathcal{U}}}

\def\gX{{\mathcal{X}}}
\def\gY{{\mathcal{Y}}}

\def\R{{\mathbb{R}}}

\newcommand{\pdata}{p_{\rm{data}}}

\newcommand{\sethree}{\mathrm{SE(3)}}

\newcommand{\pdelta}{\mathbf{\Delta}}

\newcommand{\cfm}[1]{$\sethree$-CFM\xspace}
\newcommand{\otcfm}[1]{$\sethree$-OT-CFM\xspace}
\newcommand{\sfm}[1]{$\sethree$-SFM\xspace}
\newcommand{\sflow}[0]{\textsc{Fisher-Flow}\xspace}

\newcommand\joey[1]{\noindent{\color{blue} {\bf \fbox{Joey}} {\it#1}}}
\newcommand\oscar[1]{\noindent{\color{blue} {\bf \fbox{Oscar}} {\it#1}}}

\newcommand\sam[1]{\noindent{\color{blue} {\bf \fbox{Sam}} {\it#1}}}

\newcommand{\ie}{\emph{i.e.,}~}
\newcommand{\eg}{\emph{e.g.,}~}

\renewcommand*{\backrefalt}[4]{%
    \ifcase #1 \footnotesize{(Not cited.)}%
    \or        \footnotesize{(Cited on page~#2)}%
    \else      \footnotesize{(Cited on pages~#2)}%
    \fi}
\newcolumntype{P}[1]{>{\centering\arraybackslash}p{#1}}

\hypersetup{
	colorlinks=true,       
	linkcolor=blue,        
	citecolor=blue,        
	filecolor=magenta,     
	urlcolor=blue         
}

\title{Fisher Flow Matching for Generative Modeling over Discrete Data}

\author{%
  Oscar Davis$^1$\thanks{Correspondence to \texttt{oscar.davis@cs.ox.ac.uk}.} \\
\And
  Samuel Kessler$^1$ \\
\And
  Mircea Petrache$^2$ \\
\And
  {\.I}smail {\.I}lkan Ceylan$^1$ \\
\And
  Michael Bronstein$^{1,3}$ \\
\And
  Avishek Joey Bose$^1$ \\
\AND
\normalfont $^1$University of Oxford, $^2$Pontificia Universidad Católica de Chile, $^3$Aithyra
}

\begin{document}

\maketitle

\begin{abstract}
\looseness=-1
Generative modeling over discrete data has recently seen numerous success stories, with applications spanning language modeling, biological sequence design, and graph-structured molecular data. The predominant generative modeling paradigm for discrete data is still autoregressive, with more recent alternatives based on diffusion or flow-matching falling short of their impressive performance in continuous data settings, such as image or video generation. In this work, we introduce \sflow, a novel flow-matching model for discrete data. \sflow takes a manifestly geometric perspective
by considering categorical distributions over discrete data as points residing on a statistical manifold equipped with its natural Riemannian metric: the \emph{Fisher-Rao metric}. As a result, we demonstrate discrete data itself can be continuously reparameterised to points on the positive orthant of the $d$-hypersphere $\mathbb{S}^d_+$, 
which allows us to define flows that map any source distribution to target in a principled manner by transporting mass along (closed-form) geodesics of $\mathbb{S}^d_+$. Furthermore, the learned flows in \sflow can be further bootstrapped by leveraging Riemannian optimal transport leading to improved training dynamics. We prove that the gradient flow induced by \sflow is optimal in reducing the forward KL divergence. We evaluate \sflow on an array of synthetic and diverse real-world benchmarks, including designing DNA Promoter, and DNA Enhancer sequences. Empirically, we find that \sflow improves over prior diffusion and flow-matching models on these benchmarks. Our code is available at \url{https://github.com/olsdavis/fisher-flow}.
\end{abstract}

\section{Introduction}
\label{sec:introduction}

\looseness=-1
The recent success of generative models operating on continuous data such as images has been a watershed moment for AI exceeding even the wildest expectations just a few years ago~\citep{AIForecast, bubeck2023sparks}. A key driver of this progress has come from substantial innovations in simulation-free generative models,  the most popular of which include diffusion~\citep{ho2020ddpm,song2020score} and flow matching methods~\citep{lipman_flow_2022,tong2023improving}, leading to a plethora of advances in image generation~\citep{betker2023improving, esser2024scaling,MidJourney}, video generation~\citep{Sora2024, bartal2024lumiere}, audio generation~\citep{radford2023robust}, and $3 \rm D$ protein structure generation~\citep{watson2023novo,bose2024se3stochastic}, to name a few. 

\looseness=-1
In contrast, analogous advancements in generative models over discrete data domains, such as language models~\citep{achiam2023gpt,team2023gemini}, have been dominated by autoregressive models~\citep{yule1971method}, which attribute a simple factorisation of probabilities over sequences. Modern autoregressive models, while impressive, have several key limitations which include the slow sequential sampling of tokens in a sequence, the assumption of a specified ordering over discrete objects, and the degradation of performance without important inference techniques such as nucleus sampling~\citep{holtzman2019curious}. It is expected that further progress will come from 
the principled equivalents of diffusion and flow-matching approaches for categorical distributions in the discrete data setting.

\looseness=-1
While appealing, one central barrier in constructing diffusion and flow matching over discrete spaces lies in designing an appropriate forward process that progressively corrupts discrete data.  This often involves the sophisticated design of transition kernels~\citep{austin2021structured,campbell2024generative,alamdari2023protein,lou2023discrete}, which hits an ideal stationary distribution---itself remaining an unclear quantity in the discrete setting. 
An alternative path to designing discrete transitions is to instead opt for a continuous relaxation of discrete data over a continuous space, which then enables the simple application of flow-matching and diffusion. Consequently, past work has relied on relaxing discrete data to points on the interior of the probability simplex
~\citep{avdeyev2023dirichlet,stark2024dirichlet}.
 \begin{wrapfigure}{r}{0.48\textwidth}
    \centering%
    \includegraphics[width=0.4\textwidth]{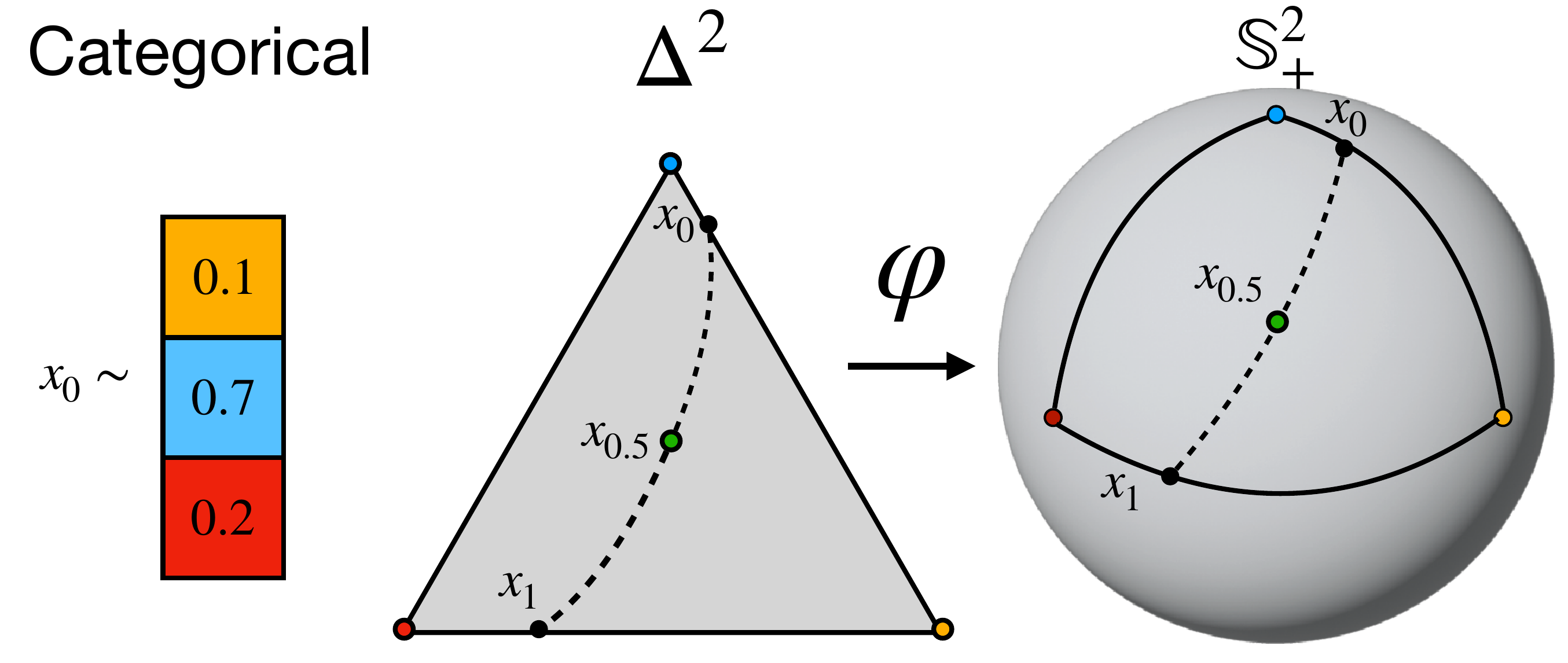}
    \caption{\small A geodesic connecting $x_0$ and $x_1$ using the FR metric on $\mathring{\Delta}^2$ and the corresponding path on $\mathbb{S}^2_+$.}
    \vspace{-5pt}
    \label{fig:visual_abstract}
\end{wrapfigure}
\looseness=-1
However,
since the probability simplex is not Euclidean, it is not possible to utilise Gaussian probability paths---the stationary distribution of an uninformative prior is uniform rather than Gaussian~\citep{chen2023riemannian}. 
One possible remedy is
to construct conditional probability paths on the simplex using Dirichlet distributions~\citep{stark2024dirichlet}, but this can lead to undesirable properties that include a complex parameterisation of the vector field. An even greater limitation is that flows using Dirichlet paths are not general enough to accommodate starting from a non-uniform (source) prior distribution---hampering downstream generative modeling applications. These limitations motivate the following research question: \emph{Can we find a continuous reparameterisation of discrete data allowing us to learn a push-forward map between any source and target distribution?}

\looseness=-1
\xhdr{Present work}
In this paper, we propose \sflow, a new flow matching-based generative model for discrete data. 
Our key geometric insight 
is to endow the probability simplex with its natural Riemannian metric---the Fisher-Rao metric---which transforms the space into a Riemannian manifold and induces a different geometry compared to past approaches of~\citet{avdeyev2023dirichlet,stark2024dirichlet}. Moreover, using this Riemannian manifold, we exploit a well-known geometric construction: the probability simplex under the Fisher-Rao metric is isometric to the positive orthant of the $d$-dimensional hypersphere $\mathbb{S}^d_+$~\citep{aastrom2017image} (see \Cref{fig:visual_abstract}). 
By operating on $\mathbb{S}^d_+$, we obtain a more flexible and numerically stable parameterisation of learned vector fields as well as the ability to use a familiar metric---namely, the Euclidean metric $\ell_2$ restricted to the sphere, which leads to better training dynamics and improved performance.
As a result, \sflow becomes an instance of Riemannian Flow Matching (RFM)~\citep{chen2023riemannian}, and our designed flows enjoy explicit and numerically favorable formulas for the trajectory connecting a pair of sampled points between \emph{any} source and target distribution---effectively generalising previous flow models~\citep{stark2024dirichlet}.

\looseness=-1
On a theoretical front, we prove in \Cref{prop:fisher_kl} that optimising the flow-matching objective with \sflow is an optimal choice for matching categorical distributions on the probability simplex. More precisely, we show the direction of the optimal induced gradient flow in the space of probabilities converges to the Fisher-Rao flow in the space of probabilities. In addition, we show in \Cref{prop:mongegeodesic} how to design straighter flows, leading to improved training dynamics, by solving the Riemannian optimal transport problem on $\posOrthant^d$. Empirically, we investigate \sflow on sequence modeling over synthetic categorical densities as well as biological sequence design tasks in DNA promoter and DNA enhancer design. We observe that our approach obtains improved performance to comparable discrete diffusion and flow matching methods of~\citet{austin2021structured,stark2024dirichlet}.

\section{Background}

\looseness=-1
The main task of generative modeling is to approximate the target distribution, $\pdata \in \gP(\gM)$, over a probability space $(\gM,\Sigma, \gP)$, using a parametric model $p_{\theta}$. The choice of  $\gM = \R^d$ appears in the classical setup of generative modeling over continuous domains, \eg images; while for categorical distributions over discrete data, we identify $\gM = \gP(\gA)$, where $\gA=\{0,\dots,d\}$ represents the categories corresponding to an alphabet with $d+1$ elements.
In this paper, we consider problem settings where the modeler has access to $\pdata$ as an empirical distribution from which the samples are drawn identically and independently. Such an empirical distribution corresponds to the {\em training} set used to train a generative model and is denoted by $\gD = \{ x_i\}_{i=1}^n$. A standard approach to generative modeling in these settings is to learn parameters $\theta$ of a generative model $p_{\theta}$ that minimises the forward KL divergence, $\mathbb{D}_{\rm KL}( \pdata || p_{\theta})$, or, in other words, maximises the log-likelihood of data under $p_{\theta}$. 

\cut{
This is equivalent to maximising the log-likelihood of observing $\pdata$, and leads to the following optimisation objective:
\begin{equation}
    \label{eq:base_obj}
    \Theta^*
    :=
    \argmax_{\theta'\in \Theta}
    \mathbb{E}_{x \sim \pdata }
    [\log p_{\theta'}(x)],
\end{equation}
\looseness=-1
where $\Theta$ represents the entire parameter space, and the set-valued $\argmax$ operator outputs the set  $\Theta^*$ of equally good solutions, and a specific solution $\theta^* \in \Theta^*$ can be selected using a tie-breaking strategy. For instance, we can simply select $\theta^*$ as the minimum norm solution $\theta^* = \argmin_{\theta' \in \Theta} \norm{\theta'}$.
}


\subsection{Information geometry}
\label{sec:information_geometry}
\looseness=-1
The space of probability distributions $\gP=\gP(X)$ over a set $X$ can be endowed with a geometric structure. 
Let $\omega$ be the parameters of a distribution such that the map $\omega \mapsto p_\omega \in \gP$ is injective. We note that this map is distinguished from the generative model, $\theta \mapsto p_{\theta}$, as $\theta$ corresponds to parameters of the \emph{neural network} rather than the parameters of the \emph{output} distribution being approximated. For instance, if we seek to model a multi-variate Gaussian $\gN(\mu, \Sigma)$ in $\R^d$, the parameters of the distribution are $\omega =(\mu, \Sigma)$, while $\theta$ can be the parameters of an arbitrary deep neural network. 

\looseness=-1
Our distributions $p_\omega$ are taken to be a family of distributions parameterised by a subset of vectors $\omega = (\omega^1, \dots, \omega^d)\in \Omega \subseteq\R^d$, with its usual topology. If the distributions $p_\omega$ are absolutely continuous w.r.t.\ a reference measure $\mu$ over $X$, with densities $p_{\omega}(x), x\in X,\omega \in\Omega$, then the injective map $\omega \in\Omega \mapsto p_{\omega} \in L^1(\mu)$ defines a \emph{statistical manifold} (cf.~\citet{amari2016information, ay2017information}):
\begin{equation}
    \gM^d := \big\{ p_{\omega}(\cdot) \, \big | \omega = (\omega^1, \dots, \omega^d) \in \Omega \subseteq \R^d \big\}. 
\end{equation}
Note that $\gM^d$ is identified as a $d$-dimensional submanifold in the space of absolutely continuous probability distributions $\gP(X)$.\footnote{
If, as in our case, we take $X=\gA=\{0,\dots,d\}$, then we can fix $\mu=\frac{1}{d+1}\sum_{i=0}^d\delta_i$, and then $\gM^d=\gP(X)$.}
If $p_\omega(x)$ is differentiable in $\omega$, then $\gM^d$ inherits a differentiable structure. We can then define a metric that converts $\gM^d$ into a \emph{Riemannian manifold}. Moreover, the parameters $\omega$ are the local coordinates and the map $\omega \mapsto p_{\omega}$ is a global parameterisation for the manifold.

\looseness=-1
As for the choice of metric, the minimisation of the forward KL divergence, $\mathbb{D}_{\rm KL}(\pdata || p_\omega)$, under mild conditions, suggests a natural prescription of a Riemannian metric on $\gM^d$~\citep{balasubramanian1996geometric}. We can arrive at this result by 
inspecting the log-likelihood of the generative model, $\log p_{\omega}$, and constructing the Fisher-information matrix whose $(i,j)$-th entry $G(\omega) = [g_{ij}(\omega)]_{ij}$ is defined as
\begin{equation}
    g_{ij}(\omega) \coloneqq \int_{\Omega}  \left(\frac{\partial \log p_{\omega}}{\partial \omega^i} \right) \left( \frac{\partial \log p_{\omega}}{\partial \omega^j} \right) p_\omega\ \diff\mu, 
\end{equation}
\looseness=-1
for $1 \leq i,j \leq d$, where $\mu$ is the reference measure on $\Omega$, which must satisfy the property that all $p_\omega$ are absolutely continuous with respect to $\mu$. In this setting, the manifestation of the Fisher-information matrix is not a mere coincidence: it is the second-order Taylor approximation of $\mathbb{D}_{\rm KL} (p_{\pdata} || p_{\omega})$ in a local neighborhood of $p_{\omega}$, in its local coordinates, $\omega$. Furthermore, the Fisher-Information matrix is symmetric and positive-definite, consequently defining a Riemannian metric. It is called the \emph{Fisher-Rao} metric and it equips a family of inner products at the tangent space $\gT_{p_{\omega}}\gM^d \times \gT_{p_{\omega}} \gM^d\to \R$ that are continuous on the statistical manifold, $\gM^d$ (they vary smoothly, in case the map $\omega\in\Omega\mapsto p_\omega\in L^1(\mu)$ is assumed to be smooth). Beyond arising as a natural consequence of KL minimisation in the generative modeling setup, the Fisher-Rao metric is the unique metric invariant to reparameterisation of $\gM^d$ (see~\citet[Thm. 1.2]{ay2017information})---a fact we later exploit in~\S\ref{sec:sphere_map} to build more scalable and more numerically stable generative models.

\subsection{Flow matching over Riemannian manifolds}
\looseness=-1
A {\em probability path} on a Riemannian manifold, $\gM^d$, is a continuous interpolation between two distributions, $p_0, p_1 \in \gP(\gM^d)$, indexed by time $t$. Let $p_t$ be a distribution on a probability path that connects $p_0$ to $p_1$ and consider its associated flow, $\psi_t$, and vector field, $u_t$. We can learn a {\em continuous normalising flow} (CNF) by directly regressing the vector field, $u_t$, with a parametric one, $v_{\theta} \in  \gT\gM^d$, where $\gT\gM^d$ is the tangent bundle. In effect, the goal of learning is to match the flow---termed \emph{flow-matching}---of the target vector field and can be formulated into a simulation-free training objective~\citep[FM]{lipman_flow_2022}, provided $p_t$ satisfies the boundary conditions, $p_0 = \pdata$ and $p_1 = p_{\text{prior}}$. As stated, the vanilla flow matching objective is intractable as we generally do not have access to the closed-form of $u_t$ that generates $p_t$. Instead, we can opt to regress $v_{\theta}$ against a conditional vector field, $u_t(x_t | z)$, generating a conditional probability path $p_t(x_t | z)$, and use it to recover the target unconditional path: $p_t(x_t) = \int_{\gM} p_t(x_t | z) q(z) \diff z$. Similarly, the vector field $u_t$ can also be recovered by marginalising conditional vector fields, $u_t (x|z)$. This allows us to state the CFM objective for Riemannian manifolds~\citep{chen2023riemannian}:
\begin{equation}\label{eq:CFM}
\gL_{\rm rcfm}(\theta) = \mathbb{E}_{t, q(z), p_t(x_t | z)} \|v_\theta(t, x_t) - u_t(x_t | z)\|_g^2, \quad t \sim \mathcal{U}(0,1).
\end{equation}
\looseness=-1
As FM and CFM objectives have the same gradients~\citep{tong2023improving, lipman_flow_2022}, at inference, we can generate by sampling from $p_1$, and using $v_{\theta}$ to propagate the ODE backwards in time. The central question in the Riemannian setting corresponds to then finding $x_t$ and $u_t(x_t | z)$. For simple geometries, one can always exploit the geodesic interpolant to construct $x_t = \exp_{x_0}(t\log_{x_0}(x_1))$ and $u_t(x_t | z) = \dot{x}_t$. 
Instead of computing the time derivative explicitly, we may also use a general closed-form expression for $u_t$, based on the geometry of the problem: $u_t=\sfrac{\log_{x_t}(x_1)}{(1-t)}$, cf.~\citep{bose2023se}.

\looseness=-1
\xhdr{Notation and convention}
We use $t \in [0,1]$ to indicate the time index of a process such that $t=0$ corresponds to $\pdata$ and $t=1$ corresponds to the terminal distribution of a (stochastic) process to be defined later. Typically, this will correspond to an easy-to-sample from source distribution. Henceforth, we use subscripts to denote the time index---\ie $p_t$---and reserve superscripts to designate indices over coordinates in a (parameter) vector, \eg $\omega^i \in ( \omega^1, \dots, \omega^d)$.
\section{Fisher Flow Matching}
\label{sec:simplex_flows}

\looseness=-1
We now establish a new methodology to perform discrete generative models under a flow-matching paradigm which we term as \sflow. Intuitively, our approach begins with the realisation that discrete data modeled as categorical distributions over $d$ categories can be parameterised
to live on the $d$-dimensional probability simplex, $\Delta^d$, whose relative interior, $\mathring{\Delta}^d$, can be identified as a Riemannian manifold endowed with the \emph{Fisher-Rao} metric~\citep{amari2016information, ay2017information,nielsen2020elementary}. Additionally, we leverage the \emph{sphere map}, which defines a diffeomorphism between the interior of the probability simplex and the positive orthant of a hypersphere, $\mathbb{S}^d_+$. As a result, generative modeling over discrete data is amenable to continuous parameterisation over spherical manifolds and offers the following key advantages: 
\begin{enumerate}[label=\textbf{(A\arabic*)},left=0pt,nosep]
\item \xhdr{Continuous reparameterisation} 
We can now seamlessly define conditional probability paths directly on the Riemannian manifold $\mathbb{S}^d_+$, enabling us to treat discrete generative modeling as continuous, through  Riemannian flow matching on the hypersphere.
\item \xhdr{Flexibility of source distribution} In stark contrast with prior work~\citep{campbell2024generative,stark2024dirichlet}, our conditional probability paths can map \emph{any} source distribution to a desired target distribution by leveraging the explicit analytic expression of the geodesics on $\mathbb{S}^d_+$.
\item \xhdr{Riemannian optimal transport} As the sphere map is an isometry of the interior of the probability simplex, we can perform Riemannian OT using the geodesic cost on $\mathbb{S}^d_+$ to construct a coupling between $p_0$ and $p_1$, leading to straighter flows and lower variance training. 
\end{enumerate}

\looseness=-1
In the following subsections, we detail first how to construct the continuous reparameterisation used in \sflow ~\S\ref{sec:continuous_reparam}. An algorithmic description of the training procedure of \sflow is presented in \Cref{alg:sflow_training}. We justify the use of the Fisher-Rao metric in~\S\ref{sec:fr_metric_theory} by showing that induces a gradient flow that minimises the KL divergence. Finally, we discuss the sphere map in~\S\ref{sec:sphere_map}, and conclude by elevating the constructed flows to minimise the Riemannian OT problem in~\S\ref{sec:riemannian_ot}.

\subsection{Reparameterising discrete data on the simplex}
\label{sec:continuous_reparam}
\looseness=-1
We now take our manifold $\gM^d = \Delta^d = \{x \in \R^{d+1} | \mathbf{1}^\top x = 1, x \geq 0\}$ as the $d$-dimensional simplex. We seek to model distributions over this space which we denote as $\gP(\Delta^d)$.
We can represent categorical distributions, $p(x)$, over $K=d+1$ categories in $\Delta^d$ by placing a Dirac $\delta_i$ with weight $p^i$ on each vertex $i \in \{0,\dots,d\}$.\footnote{We denote, with a slight abuse of notation, the probability of category $i$ by $p^i$, \ie $\sum_i p^i =1$.} Thus a discrete probability distribution given by a categorical can be converted into a continuous representation over $\Delta^d$ by representing the categories $p^i$ as a mixture of point masses at each vertex of $\Delta^d$. This allows us to write our data distribution $\pdata$ over discrete objects as:
\begin{equation}
    \pdata(x) = \sum_{i=0}^d p^i \delta(x - e_i),
    \label{eqn:dirac_on_vertex}
\end{equation}
\looseness=-1
where $e_i$ are $K=d+1$ one-hot vectors representing the vertices of the probability simplex\footnote{Note that $e_i\in \Delta^d$ represents Dirac mass $\delta_i\in\gP(\gA)$, thus \eqref{eqn:dirac_on_vertex} means that $\pdata =\sum_i p^i \delta_{\delta_i}\in\gP(\gP(\gA))\simeq\gP(\Delta^d)$. The traditional form $\sum_ip^i\delta_i\in \gP(\gA)$ is recovered via the identification $\delta_i\mapsto i$.}. 
While the vertices of $\Delta^d$ are still discrete, the relative interior of the probability simplex, denoted as $\mathring{\Delta}^d:=\{x\in\Delta^d:\ x>0\}$, is a continuous space, whose geometry can be leveraged to build our method, \sflow. 
Consequently, we may move Dirac masses on the vertices of the probability simplex to its interior---and thereby performing continuous reparameterisation ---by simply applying any smoothing function $\sigma: \Delta^d\to \mathring{\Delta}^d$, \eg label smoothing as in supervised learning~\citep{szegedy2016rethinking}.

\looseness=-1
\xhdr{Defining a Riemannian metric} Relaxing categorical distributions to the relative interior, $\mathring{\Delta}^d$, enables us to consider a more geometric approach to building generative models. Specifically, this construction necessitates that we treat $\mathring{\Delta}^d$ as a \emph{statistical Riemannian manifold} wherein the geometry of the problem corresponds to classical \emph{information geometry}~\citep{amari2016information,ay2017information, nielsen2020elementary}. This leads to a natural choice of Riemannian metric: the Fisher-Rao metric, defined as, on the tangent space at an interior point $p \in \mathring{\Delta}^d$,
\begin{equation}
    \forall u, v \in \gT_p \mathring{\Delta}^d, \quad g_{\rm FR}(p)[u,v] \coloneqq \langle u, v \rangle_p := \left \langle \frac{u}{\sqrt{p}}, \frac{v}{\sqrt{p}}\right \rangle_2 = \sum_{i=0}^d \frac{u^i v^i}{p^i}.
    \label{eqn:fisher_rao_metric}
\end{equation}
\looseness=-1
In the above equation, the inner product normalisation by $\sqrt{p}$ is applied component-wise. After normalising by $\sqrt{p}$ the inner product on the simplex becomes synonymous with the familiar Euclidean inner product $\langle \cdot, \cdot \rangle_2$. However, near the boundary of the simplex, this ``tautological'' parameterisation of the metric by the components of $p$ is numerically unstable due to division by zero. This motivates a search for a more numerically stable solution which we find through the sphere-map in~\S\ref{sec:sphere_map}.
As a Riemannian metric is a choice of inner product that varies smoothly, it can be readily used to define geometric quantities of interest such as distances between points or angles, as well as a metric-induced norm. We refer the interested reader to~\S\ref{app:geometry_simplex} for more details on the geometry of $\mathring{\Delta}^d$.

\subsection{Flow Matching from $\mathring{\Delta}^d \to \mathbb{S}^d_+$ via the sphere map}
\label{sec:sphere_map}

\looseness=-1
The continuous parameterisation of categorical distributions to the interior of the probability simplex, while theoretically appealing, can be prone to numerical challenges. This is primarily because in practice we do not have the explicit probabilities of the input distribution, but instead, one-hot encoded samples which 
means that we must flow to a vertex. More concretely,  this implies that when $t \to 1$, $x_t \to e_i$ for some $i \in [d]$, therefore implying $\norm{v}_{x_t}\to \infty$, where $\norm{\cdot}_{x_t}$ denotes the norm at point $x_t$. This occurs due to the metric normalisation $\sqrt{p}$, applied component-wise. In addition, the restriction of $v_{\theta}$ to be at the tangent space imposes architectural constraints on the network. What we instead seek is a flow parameterisation without any architectural restrictions or numerical instability due to the metric norm. We achieve this through the \emph{sphere map}, $\varphi: \mathring\Delta^d \to \mathbb{S}^d_+$, which is a diffeomorphism between the interior of the simplex and an open subset of the positive orthant of a $d$-dimensional hypersphere.
\begin{equation}
    \begin{aligned}
        \varphi: \mathring\Delta^d \longrightarrow \mathbb{S}^d_+, \quad p \longmapsto s := \varphi(p) =  \sqrt{p},\\
    \varphi^{-1}: \mathbb{S}^d_+\longrightarrow \mathring\Delta^d, \quad s \longmapsto p :=\varphi^{-1}(s) = s^2.
    \end{aligned}
    \label{eqn:sphere_map_and_inverse}
\end{equation}
\looseness=-1
In \eqref{eqn:sphere_map_and_inverse}, both the sphere map and its inverse are operations that are applied element-wise. The sphere map reparameterisation identifies the Fisher-Rao geometry of $\mathring\Delta^d$ to the geometry of a hypersphere, whose Riemannian metric is induced by the Euclidean inner product of $\R^{d+1}$. It is easy to show that $2\varphi$ (the sphere map scaled by $2$) preserves the Riemannian metric of $\mathring\Delta^d$, \ie it is an isometry, and that therefore all geometric notions such as distances are also preserved. However, a key benefit we obtain is that we can \emph{extend} the metric to the boundary of the manifold without introducing numerical instability as the metric at the boundary does not require us to divide by zero.


\looseness=-1
\xhdr{Building conditional paths and vector fields on $\mathbb{S}^d_+$}
On any Riemannian manifold $\gM^d$ that admits a probability density, it is possible to define a geodesic interpolant that connects two points between samples $x_0 \sim p_0$ to $x_1 \sim p_1$. A point traversing this interpolant, indexed by time $t \in [0,1]$, can be expressed as $x_t = \exp_{x_0}(t\log_{x_0}(x_1))$. On general Riemannian manifolds, it is often not possible to obtain analytic expressions for the manifold exponential and logarithmic maps and as a result, traversing this interpolant requires the costly simulation of the Euler-Lagrange equations. Conveniently, under the Fisher-Rao metric $\mathring{\Delta}^d$ admits simple analytic expressions for the exponential and logarithmic maps---and consequently the geodesic interpolant. Moreover, due to the sphere-map $\varphi$ in \cref{eqn:sphere_map_and_inverse} the geodesic interpolant is also well-defined on $\posOrthant^d$. Such a result means that the conditional flow $x_t$ on $\posOrthant^d$ can be derived analytically from the well-known geodesics on a hypersphere, \ie they are the great circles but restricted to the positive orthant. Consequently, we may build all of the conditional flow machinery using well-studied geometric expressions for $\mathbb{S}^d_+$ in a numerically stable manner. 

\looseness=-1
The target conditional vector field associated at $x_t$ can also be written in closed-form $u_t(x_t | x_0, x_1) = \log_{x_t}(x_1) / (1-t)$ and computed exactly on $\posOrthant^d$. Intuitively, $u_t$ moves at constant velocity from $x_t$ in the direction of $x_1$ and presents a simple regression target to learn the vector field $v_{\theta}$. 
One practical benefit of learning conditional vector fields on $\posOrthant^d$ is that it allows for more flexible parameterisation of the vector field network $v_{\theta}$. Specifically,
the network $v_{\theta}$ can be unconstrained and output directly in the ambient space $\R^{d+1}$ after which we can orthogonally project them to the tangent space of $x_t$. This is possible since we can take an \emph{extrinsic} view on the geometry and isometrically embed $\posOrthant^d$ to the higher dimensional ambient space due to the Nash embedding theorem~\citep{gunther1991isometric}. More formally, we have that $v_{\theta}(t, x_t) = \phi_{x_t}(\tilde{v}_{\theta}(t, x_t))$, where $\tilde{v}_{\theta}$ is the output in $\R^d$ and $\phi_{x_t}: \R^d\to\tangent_{x_t}\posOrthant^d$ and is defined as $ \phi_{x_t}(\tilde{v}) = \tilde{v} - \langle x_t, \tilde{v} \rangle_2 x_t$.

\looseness=-1
In the absence of any knowledge we can choose an uninformative prior on $\posOrthant^d$ which is the uniform density over the manifold $p_1 (x_1) = \sqrt{\det \mathbf{G}(x_1)} / \int_{\posOrthant^d} \sqrt{\det \mathbf{G}(x_1)}$, where $\mathbf{G}$ is the matrix representation of the Riemannian metric.
However, a key asset of our construction, in contrast, to~\citep{campbell2024generative,stark2024dirichlet}, is that $p_1$ can be any source distribution since we operate on the \emph{interpolant-level} by building geodesics between two points, $x_0, x_1 \in \posOrthant^d$. We now state the Riemannian CFM objective for $\posOrthant^d$:
\begin{equation}\label{eq:Simplex_CFM}
\gL_{\mathbb{S}^d_+}(\theta) = \mathbb{E}_{t, q(x_0, x_1), p_t(x_t | x_0, x_1)} \|v_\theta(t, x_t) - \log_{x_t}(x_1)/(1-t)\|_{\mathbb{S}^d_+}^2, \quad t \sim \mathcal{U}(0,1).
\end{equation}

\looseness=-1
In a nutshell, our recipe for learning conditional flow matching for discrete data first maps the input data to $\posOrthant^d$. Then we learn to regress target conditional vector fields on $\posOrthant^d$ by performing Riemannian CFM which can be done easily as the hypersphere is a simple geometric object where geodesics can be stated explicitly. At inference, our flow pushes forward a prior on $p_1 \in \posOrthant^d$ to a desired target, $p_0$, which is then finally mapped back to $\mathring\Delta^d$ using $\varphi^{-1}$. A discrete category can then be chosen using any decoding strategy such as sampling using the mapped categorical or greedily by simply selecting the closest vertex of the probability simplex $\Delta^d$ to the final point at the end of inference.



\subsection{The Fisher-Rao metric from Natural gradient descent}
\label{sec:fr_metric_theory}
\looseness=-1
We now motivate the choice of the Fisher-Rao metric as not only a natural choice but also the optimal one on the probability simplex. We show that gradient descent of the general form $\delta\theta\mapsto\argmin_{|\delta\theta|\leq \epsilon}\mathcal L(\theta+\delta\theta)$ (for $\mathcal L(\theta)=\mathcal L(p_\theta)$ as in \eqref{eq:Simplex_CFM}) converges to the gradient flow (of parameterised probabilities $p_\theta$, or of probability paths $p_{\theta,t}$) with respect to the Wasserstein distance on $\gP(\gM)$ induced by Fisher-Rao metric $g_{\rm{FR}}$ over $\gM$. Equivalently, we get the canonical metric over $\posOrthant^d$ due to the isometry. This presents a further justification for the use of the Fisher-Rao metric. 

\looseness=-1
In order to present the gradient flow of $\mathcal L:\gP(\gM^d)\to\R$ in which $(\gM^d,g)$ is a Riemannian manifold, we recall the basics of geometry over probability spaces~\citep{ambrosio2005gradient, villani2009optimal}. If $d_g$ is the geodesic distance associated to $g$ then $W_{2,g}$ will be the optimal transport distance over $\gP=\gP(\gM^d)$ with cost $d_g^2(x,y)$. Then $(\gP(\gM^d),W_{2,g})$ is an infinite-dimensional Riemannian manifold, in which for $p\in\gP(\gM^d)$ we have the tangent space $\gT_p\gP\simeq \overline{\{\nabla_g \phi:\ \phi\in C^1_c(\gM^d)\}}^{L^2_g(\gT\gM^d; p)}$, \ie the closure of gradient vector fields with respect to $L^2_g(\gT\gM^d; p)$-norm. This norm is defined by the Riemannian tensor $g^{\gP}$ induced by $g$, which at $v,w\in \gT_p\gP$ is given by $g^{\gP}(v,w):=\int_{\gM^d}\langle v(x), w(x)\rangle_g\ dp(x)$ . In particular, note that a choice of Riemannian metric $g$ over $\gM^d$ specifies a unique metric $g^\gP$ over $\gP$. 

In the following, at the onset, we assume a bounded metric, $g$, over $\Delta^d$, which we use only to state our Lipschitz dependence assumptions. If we compare categorical densities (elements of $\gM^d=\gP(\gA)$ via KL-divergence, then it is natural to compare distributions $\mu,\nu\in \gP(\gM^d)$ via the Wasserstein-like  $W_{\rm KL}(\mu,\nu):=\min_{\pi\in \Pi(\mu,\nu)}\mathbb E_{(p_\omega, p_{\omega'})\sim \pi}[\mathbb{D}_{\rm KL}(p_\omega||p_{\omega'})]$. In the next proposition, we show that the Fisher-Rao metric appears naturally in the continuum limit of our gradient descent over $\gP(\gM^d)$. 


\begin{mdframed}[style=MyFrame2]
\begin{restatable}{proposition}{fisherkl}
\label{prop:fisher_kl}
    Assume that there exists a bounded Riemannian metric $g$ over $\Delta^d$ such that the parameterisation map $\theta\mapsto p=p(\theta)$ is Lipschitz and differentiable from $\Theta$ to $(\gP(\gM), W_{2,g})$. Then the "natural gradient" descent of the form:
    \begin{equation}\label{eq:graddesc}
    p(\theta_{n+1})\in \argmin\left\{\mathcal L(p(\theta_{n+1})):\ W_{\rm KL}(p(\theta_{n+1}), p(\theta_n))\leq\epsilon\right\}
    \end{equation}
    approximates, as $\epsilon\to 0^+$, the gradient flow of $\mathcal L$ on manifold $(\gP(\gM^d), W_{g_{\rm{FR}},2})$ with metric $g_{\rm{FR}}^\gP$ induced by Fisher-Rao metric $g_{\rm{FR}}$:\looseness=-1
    \begin{equation}\label{eq:gradflow}
        \tfrac{\diff}{\diff s}p(\theta(s)) = \nabla_{g_{\rm{FR}}^{\gP}}\mathcal L(p(\theta(s))).
    \end{equation}
\end{restatable}
\end{mdframed}
\cut{
\begin{proofsketch}
\looseness=-1
    We present the full proof~\S\ref{app:frjustif} and a simple proof sketch here. Given the assumption in~\cref{prop:fisher_kl} we can parameterise the update, $p(\theta)\mapsto p(\theta+\delta\theta)$, by a transport map $T$ under which each point $p_\omega\in \gM$ moves a uniformly bounded $\mathbb{D}_{\rm KL}$-distance,
\begin{equation*}\label{eq:natgrad}
    p(\theta+\delta\theta)\in \argmin\left\{\mathcal L(p'):\ \exists T:\gM^d\to\gM^d, p'=T_\#p(\theta),\ \mathbb D_{\rm{KL}}(p_\omega||T(p_\omega))\leq C\epsilon\right\}.
\end{equation*}
Then the approximation of $T_\#p(\theta)$ as the flow of $p(\theta)$ under the vector field $v(p_\omega):=\log_{p_\omega}(T(p_\omega))$ and the approximation of $\mathbb{D}_{\rm KL}(p_\omega||T(p_\omega))\simeq \frac12 g_{\rm{FR}}(v(p_\omega),v(p_\omega))$ allow us to conclude.
\end{proofsketch}
}
\looseness=-1
For the proof, see~\S\ref{app:frjustif}. We distinguish the results of \Cref{prop:fisher_kl} from those of Natural Gradients used in classical NN optimisation such as KFAC~\citep{martens2015optimizing}.
Note that in regular NN training, Natural Gradients~\citep{pascanu2013revisiting} implement a second-order optimisation to tame the gradient descent, at a nontrivial computational cost. Thus, the above proposition implies that, just by selecting $g_{\rm{FR}}$ metric over $\Delta^d$, we directly get the benefits that are equivalent to the regularisation procedure of Natural Gradient.

\subsection{\sflow Matching with Riemannian optimal transport}
\label{sec:riemannian_ot}

\looseness=-1
We now demonstrate how to build conditional flows that minimise a Riemannian optimal transport (OT) cost. Flows constructed by following an optimal transport plan enjoy several theoretical and practical benefits: 1. They lead to shorter global paths. 2. No two paths cross which leads to lower variance gradients during training. 3. Paths that follow the transport plan have lower kinetic energy which often leads to improved empirical performance due to improved training dynamics~\citep{shaul2023kinetic}.

\looseness=-1
The Riemannian optimal transport for \sflow can be stated for either $\mathring{\Delta}^d$ under the Fisher-Rao metric or $\posOrthant^d$. Both instantiations lead to the same optimal plan due to the isometry between the two manifolds. Specifically, we couple $q(x_0), q(x_1)$ via the Optimal Transport (OT) plan $\pi(x_0, x_1)$ under square-distance cost $c(x,y):=d^2(x,y)$---\ie $\pi(x_0, x_1)$ will be the minimiser of $\mathbb E_{(x_0,x_1)\sim \pi'}[d^2(x_0,x_1)]$ amongst all couplings $\pi'$ of fixed marginals $q(x_0), q(x_1)$. 
Now, recall that Wasserstein distance $W_2$ over $\gP(\posOrthant^d)$ is defined as $W_2(\mu,\nu):=\min_{\pi'}\mathbb E_{(x,y)\sim \pi'}[d^2_{\posOrthant^d}(x,y)]$, in which the minimisation is amongst transport plans from $\mu$ to $\nu$, defined as probability measures over $\posOrthant^d\times\posOrthant^d$ whose two marginals are respectively $\mu$ and $\nu$ \citep{villani2003topics}. Since $\posOrthant^d$ is a smooth bounded uniquely geodesic Riemannian manifold with boundary, the metric space $(\gP(\posOrthant^d), W_2)$ is uniquely geodesic and we have the following ``informal'' proposition (see \S\ref{app:ot} for the full statement):

\begin{mdframed}[style=MyFrame2]
\begin{restatable}{proposition}{mongegeodesic}
\label{prop:mongegeodesic}
    For any two Borel probability measures $p_0, p_1\in\mathcal P(\posOrthant^d)$, there exists a unique OT-plan $\pi$ between $p_0,p_1$. If $e_t(x_0,x_1)$ is the constant-speed parameterisation of the unique geodesic of extremes $x_0$ and $x_1$, and $e_t:\posOrthant^d\times\posOrthant^d\to\posOrthant^d$ is given by $ e_t(x_0,x_1)\coloneqq\exp_{x_0}(t\log_{x_0}(x_1))$, then there exists a unique Wasserstein geodesic $(p_t)_{t\in[0,1]}$ connecting $p_0$ to $p_1$, given by
    \begin{equation}\label{eq:displconv}
        p_t\coloneqq(e_t)_\#\pi\in \gP(\posOrthant^d),\quad t\in [0,1].
    \end{equation}
\end{restatable}
\end{mdframed}
\looseness=-1
The complete statement of \Cref{prop:mongegeodesicapp} along with its proof is provided in~\S\ref{app:ot}. As a consequence of \Cref{prop:mongegeodesic} we use the Wasserstein geodesic as our target conditional probability path. Operationally, this requires us to sample from marginals $x_0 \sim p_0$ and $x_1 \sim p_1$ and solve for the OT plan $\pi$ using the squared distance on $\posOrthant^d$ as the cost which is done using the Sinkhorn algorithm~\citep{genevay2018learning}.

\subsection{Training \sflow}
\label{sec:training}

\looseness=-1
\xhdr{Generalising to sequences}
Many problems in generative modeling over discrete data are concerned with handling a set or a sequence of discrete objects. For complete generality, we now extend \sflow to a sequence of discrete data by modeling it as a Cartesian product of categorical distributions. Formally, for a sequence of length $k$ we have a distribution over a product manifold $\gP(\pdelta):= \gP(\Delta^d_1) \times \dots \times \gP(\Delta^d_k)$. Equipping each manifold in the product with the Fisher-Rao metric allows us to extend the metric in a natural way to $\pdelta$. Moreover, by invoking the diffeomorphism using the sphere-map $\varphi$ independently we achieve the product of $d$-hyperspheres restricted to the positive orthant. Stated explicitly, a sequence of categorical distributions is $\gP(\mathbf{S_+}):= \gP((\posOrthant^d)_1) \times \dots \times \gP((\posOrthant^d)_k)$. Due to the factorisation of the metric across the product space, we can build independent flows on each manifold $\posOrthant^d$ and couple them in a natural way using the product metric to induce a flow on $\mathbf{S_+}$.

\looseness=-1
\xhdr{Training}
We detail our method for training \sflow in \Cref{alg:sflow_training} in~\S\ref{app:algorithm_box}. Training \sflow requires two input distributions: a source and a target one. In the case of unconditional generation, one can take $p_0 = \gU(\posOrthant^d)$, by default. In some settings, it is possible to incorporate additional conditional information, $c$. This can easily be accommodated in \sflow by directly inputting this into the parameterised vector field network with $v_\theta(\cdot)$ becoming $v_\theta(\cdot|c)$.

\cut{
\begin{algorithm}[t]
\caption{\sflow, training on $\posOrthant^d$.}
\label{alg:sflow_training}
\begin{algorithmic}[1]
\State \textbf{Input:} Source and target distributions, $\rho_1, \rho_0$, flow network $v_\theta$.
\While{Training}
    \State $t, x_0, x_1 \sim \gU(0, 1), p_0, p_1 = p_{\mathrm{data}}$
    \State $\bar{\pi} \gets \text{OT}_{\posOrthant^d}(x_0, x_1)$ \Comment{Since $x_1$ is one-hot encoded, it is on $\posOrthant^d$.}
    \State $r_0, r_1 \sim \bar{\pi}$
    \State $r_t \gets \exp_{r_0}(t \log_{r_0}(r_1))$\Comment{Geodesic interpolant between $r_0, r_1 \in \posOrthant^d$.}
    \State $u_t(r_t|r_0, r_1) \gets \dot{r}_t$\Comment{Calculated either explicitly or with a numerical approximation.}
    \State $\gL_{\sflow} \gets \norm*{v_\theta(t, r_t) - u_t(r_t|r_0, r_1)}_{\posOrthant^d}^2$
    \State $\theta \gets \text{Update}(\theta, \nabla_\theta \gL_{\sflow})$
\EndWhile
\State \textbf{return} $v_\theta$
\end{algorithmic}

\end{algorithm}

}

\section{Experiments}
\label{sec:experiments}

\looseness=-1
We now investigate the empirical caliber of \sflow on a range of synthetic and real-world benchmarks outlined below. Unless stated otherwise, all instantiations of \sflow use the optimal transport coupling on minibatches. Exact implementation details are included in~\S\ref{app:implementation_details}.

\subsection{Synthetic experiments}
\label{sec:synth_experiments}
\begin{figure}[tb]
    \vspace{-10pt}
    \centering
    \begin{subfigure}[t]{0.19\textwidth}
    \centering
    \includegraphics[width=1\textwidth]{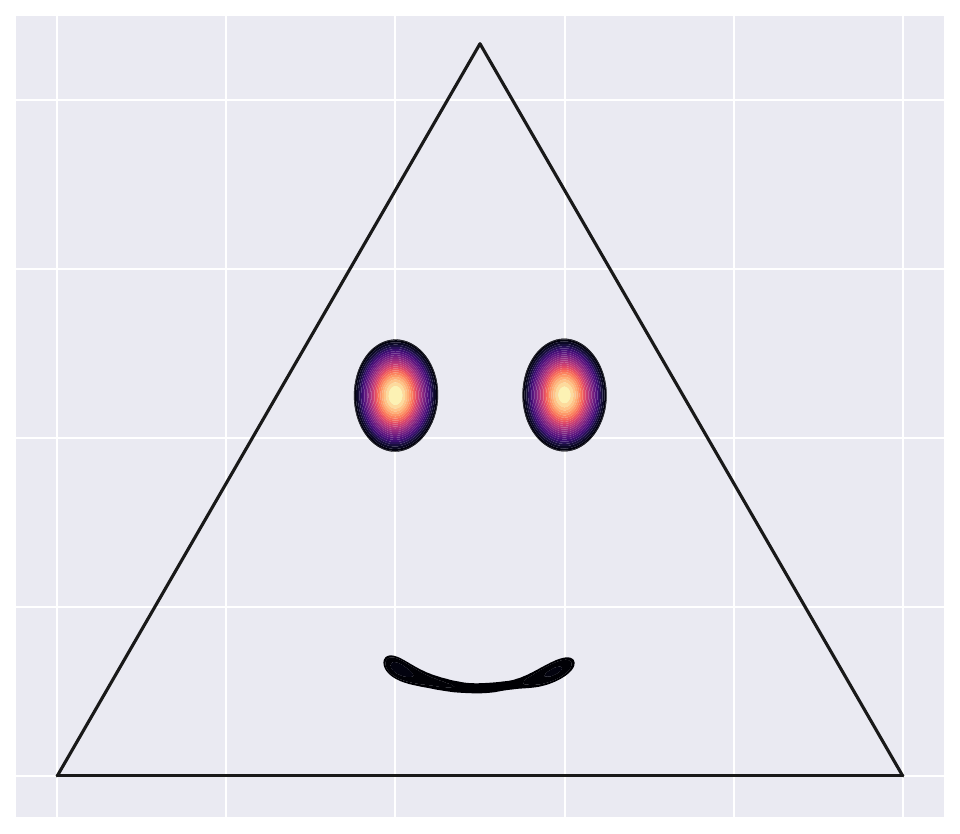}
    \caption{Real distribution}
    \end{subfigure}
    \begin{subfigure}[t]{0.19\textwidth}
    \centering
    \includegraphics[width=1\textwidth]{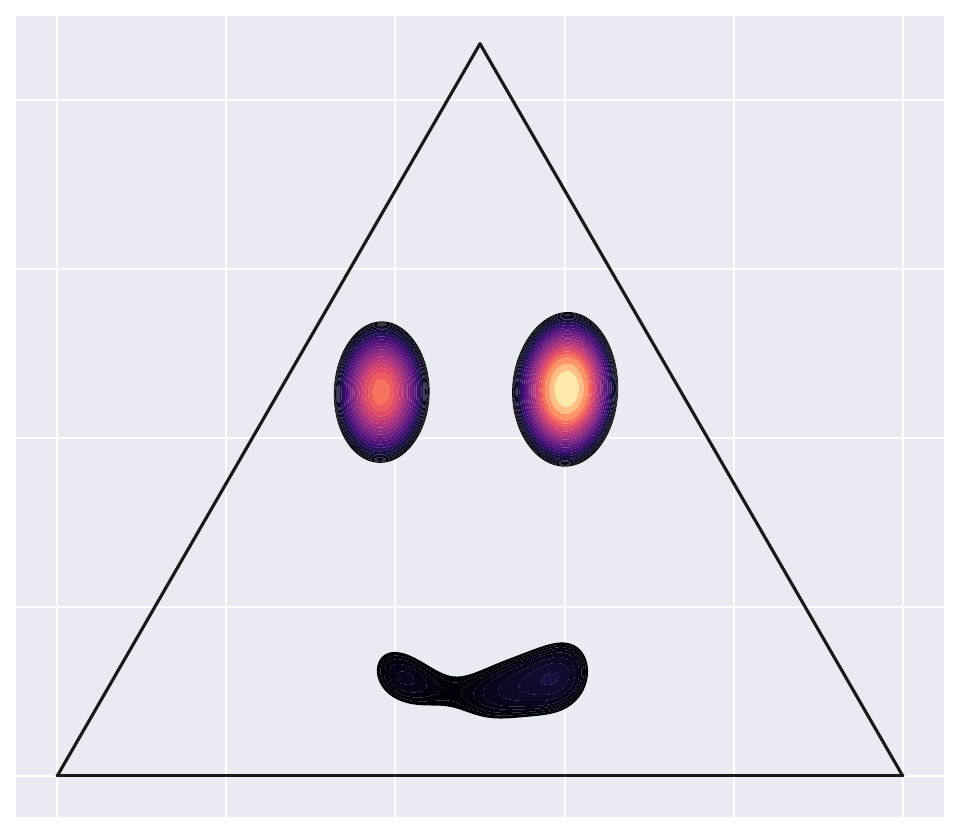}
    \caption{FF on $\mathring{\Delta}^2$}
    \end{subfigure}
    \begin{subfigure}[t]{0.19\textwidth}
    \centering
    \includegraphics[width=1\textwidth]{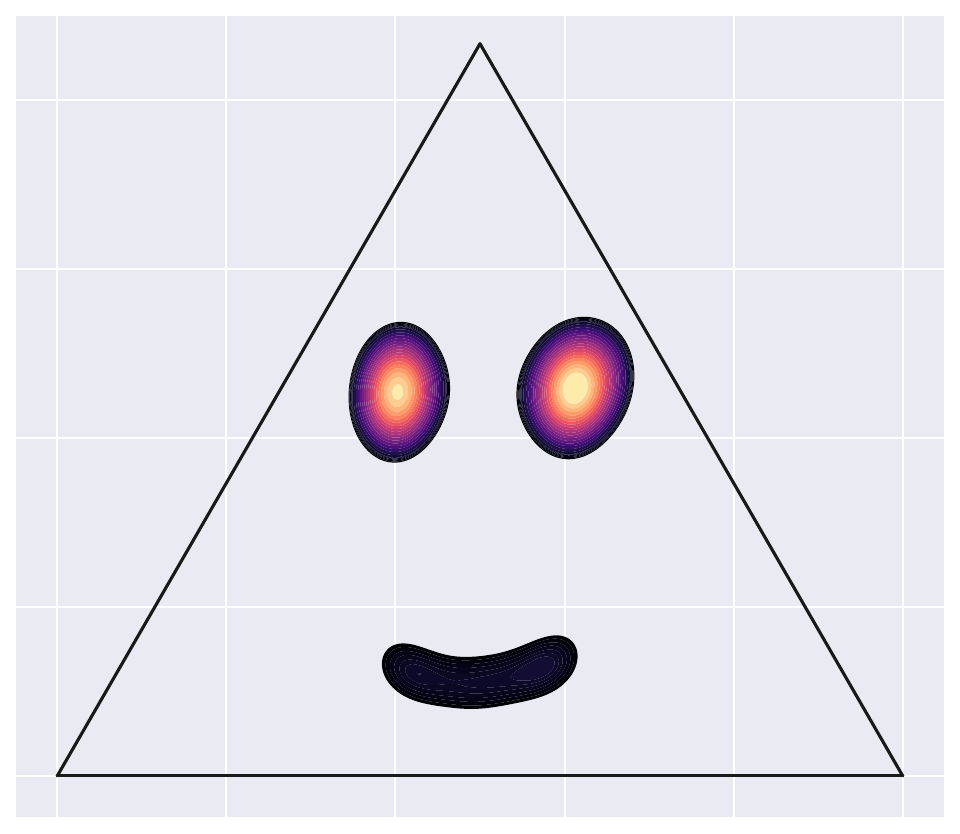}
    \caption{FF-OT on $\mathring{\Delta}^2$}
    \end{subfigure}
    \begin{subfigure}[t]{0.19\textwidth}
    \centering
    \includegraphics[width=1\textwidth]{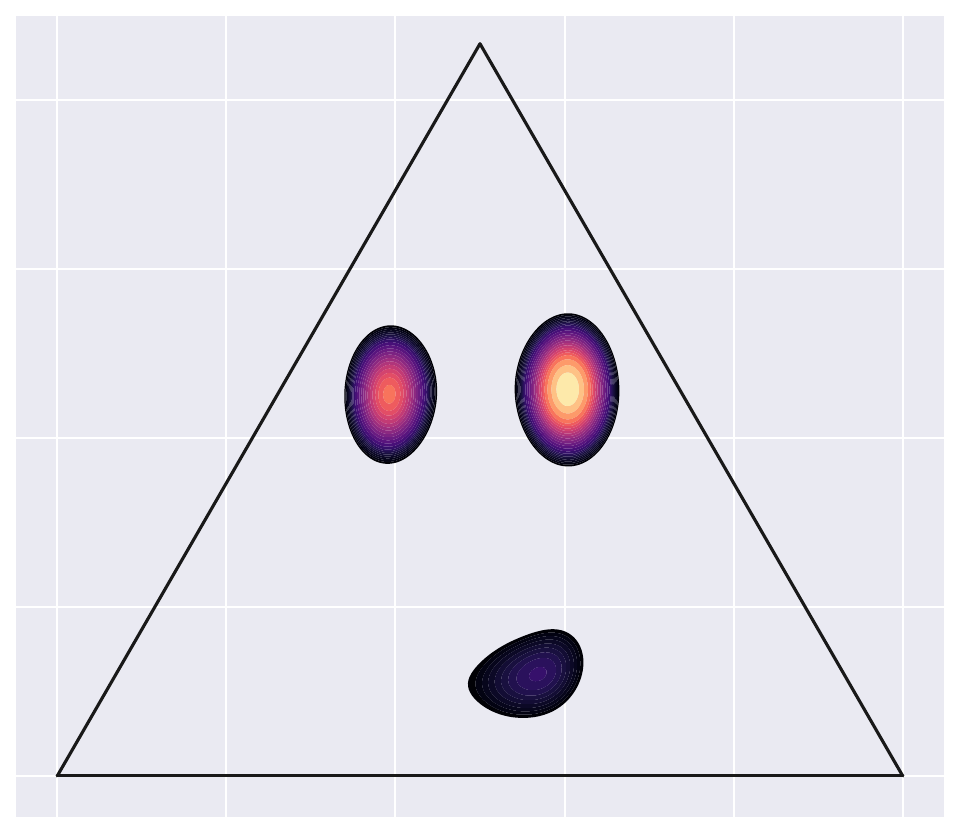}
    \caption{FF on $\posOrthant^2$}
    \end{subfigure}
    \begin{subfigure}[t]{0.19\textwidth}
    \centering
    \includegraphics[width=1\textwidth]{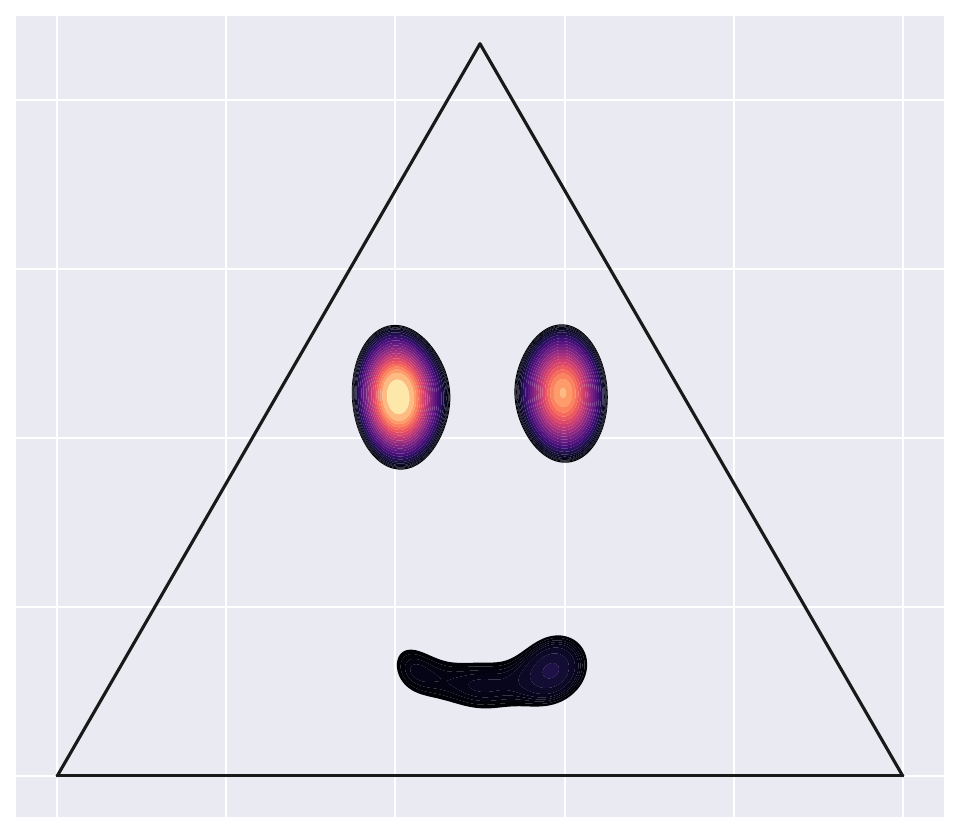}
    \caption{FF-OT on $\posOrthant^2$}
    \end{subfigure}
    \vspace{-5pt}
    \caption{\small Synthetic experiments on learning a distribution resembling a smiley face on $\mathring{\Delta^2}$. }
    \vspace{-5pt}
    \label{fig:smiley_experiment}
\end{figure}

\looseness=-1
\xhdr{Density estimation} In this first experiment, we model an empirical categorical distribution visualized on $\mathring{\Delta}^2$. In \Cref{fig:smiley_experiment}, we observe that \sflow instantiated on $\posOrthant^2$ with OT is the best in modeling the ground truth distribution. Both learning on the simplex and the positive orthant benefit from OT.

\begin{figure}
    \centering
    \begin{subfigure}[t]{0.45\textwidth}
        \includegraphics[width=1\textwidth]{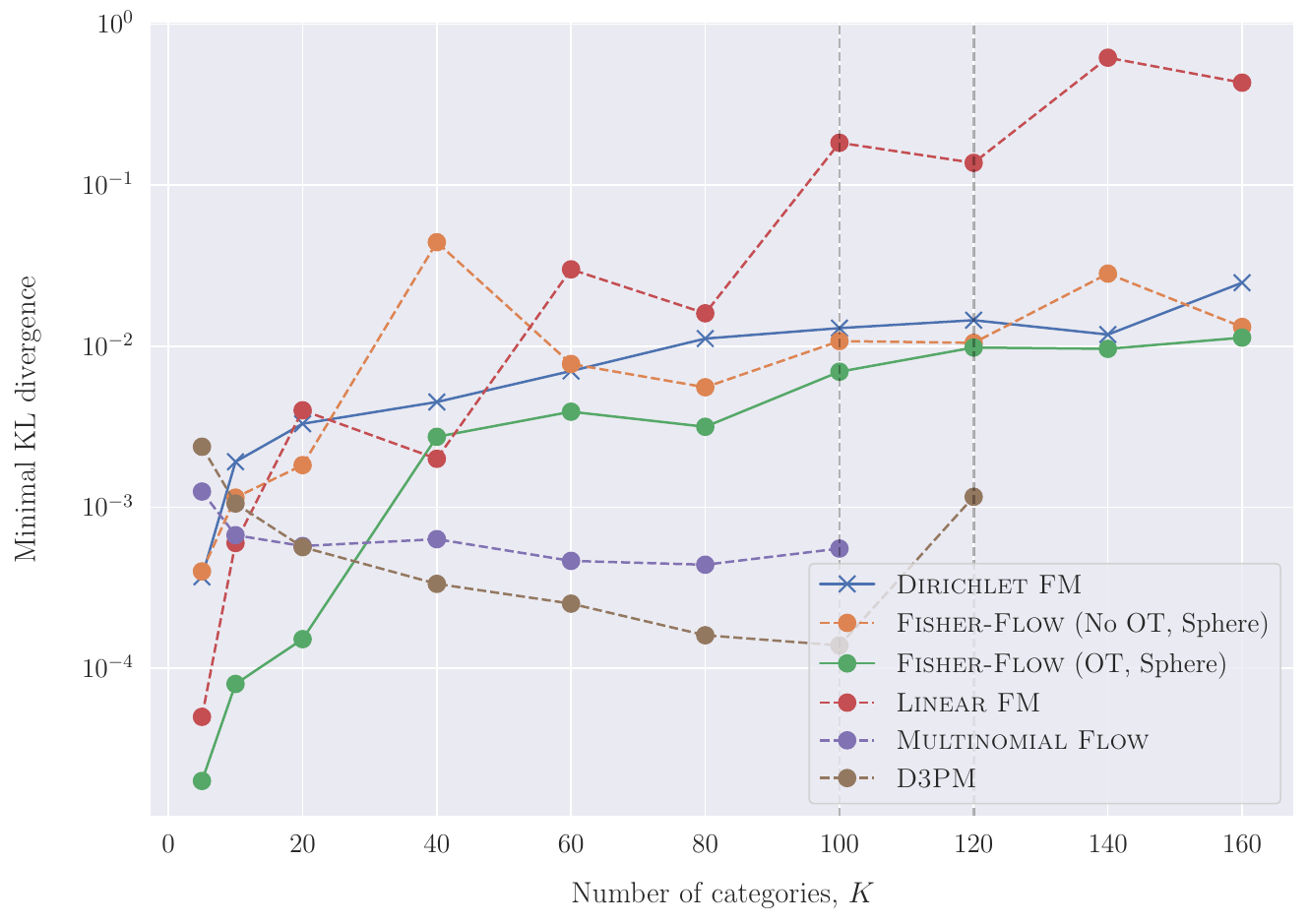}
        \caption{Results of the ablation study. Missing points for Multinomial Flow \citep{hoogeboom2021multinomialflow} and D3PM \citep{austin2021structured} are \texttt{NaN}s.}
        \label{fig:dfm_ablation}
    \end{subfigure}
    \hfill
    \begin{subfigure}[t]{0.45\textwidth}
        \includegraphics[width=1\textwidth]{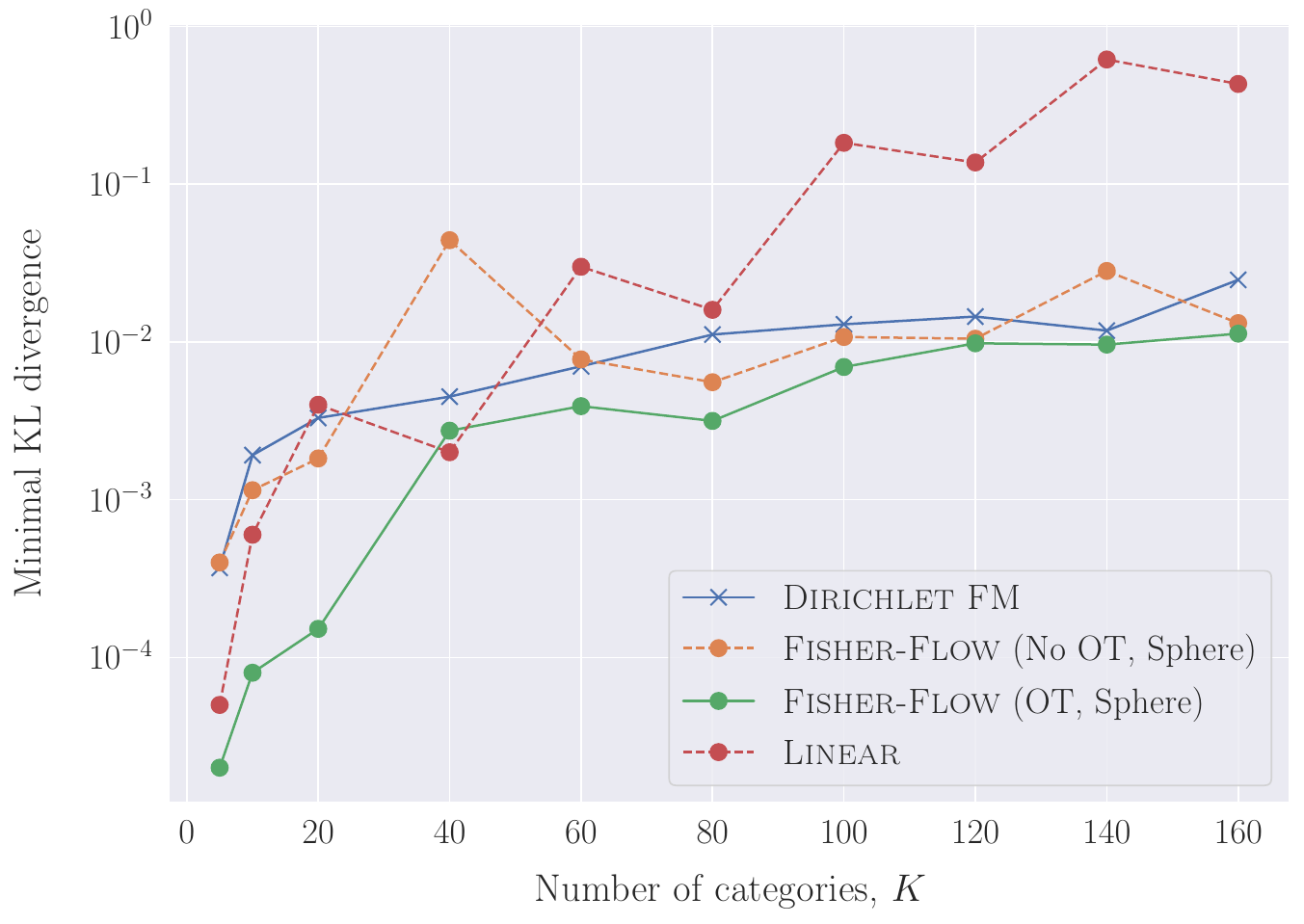}
        \caption{KL divergence of \sflow against {\scshape Dirichlet FM}.}
        \label{fig:dfm_kl_vs}
    \end{subfigure}
    \caption{Toy experiment from \citet{stark2024dirichlet}. Minimal KL divergence over 5 seeds is reported.}
    \label{fig:dfm_synthetic_experiment}
    \vspace{-10pt}
\end{figure}

\looseness=-1
\xhdr{Density learning in arbitrary dimensions} We also consider the toy experiment of \citet{stark2024dirichlet}, where we seek to model a random distribution over $(\Delta^K)^4$ for $K \in \mathbb{N}^\star$. TWhe KL divergence between the estimated distribution over $512{,}000$ samples and the true generated distribution is used as the evaluation metric. Details are provided in \S\ref{app:toy_experiment}. Results in~\Cref{fig:dfm_kl_vs} demonstrate that \sflow outperforms {\scshape Dirichlet FM}, while remaining competitive against D3PM \citep{austin2021structured} and Multinomial Flow \citep{hoogeboom2021multinomialflow}, especially in high dimensions, in which both exhibit unstable behaviour. We also conduct an ablation in \Cref{fig:dfm_ablation} and find that using optimal transport helps for both \sflow on $\mathring{\Delta}^d$ and $\posOrthant^d$, with the latter leading to the best performance.

\subsection{Promoter DNA sequence design}
\label{sec:dna_promoter}
\begin{table}[htb]
    \parbox[t]{.45\linewidth}{
    \centering
    \caption{\small MSE of the transcription profile conditioned on generated promoter DNA sequences over the test set. The last $3$ MSE and PPL values are from $5$ independent experiments. The remaining numbers are taken directly from~\citet{stark2024dirichlet}.}
    \resizebox{0.45\textwidth}{!}{
    \begin{tabular}{@{}lcc@{}} 
        \toprule
        \textbf{Model} & \textbf{MSE} ($\downarrow$) &  \textbf{PPL} ($\downarrow$) \\
        \midrule
        {\scshape Bit Diffusion (bit-encoding)} & $0.041$ & --- \\
        {\scshape Bit Diffusion (one-hot encoding)} & $0.040$ & --- \\
        {\scshape D3PM-uniform} & $0.038$ & ---\\
        {\scshape DDSM} & $0.033$ & ---\\
        {\scshape Language Model} & $0.034 \pm 0.001$ & $2.247 \pm 0.102$\\
        {\scshape Dirichlet FM} & $0.034 \pm 0.004$ & $\mathbf{1.978 \pm 0.006}$\\ 
        \midrule
        \sflow (ours) & $\mathbf{0.029\pm 0.001}$ & $\mathbf{1.4\pm2.7}$\\ 
        \bottomrule
    \end{tabular}
    }
    \label{tab:promoter}
    }
    \hfill
    \parbox[t]{.45\linewidth}{
    \centering\caption{\small Perplexities (PPL) values for different methods for enhancer DNA generation. Lower PPL is better. Values are an average and standard error over $5$ seeds.}
\resizebox{0.45\textwidth}{!}{%
\begin{tabular}{@{}lcc@{}}
\toprule
\textbf{Method}           & \textbf{Melanoma PPL} ($\downarrow$)   & \textbf{Fly Brain PPL} ($\downarrow$) \\ 
\midrule
Random Sequence   & $895.88$ & $895.88$ \\
Language Model    & $2.22 \pm 0.09$  & $2.19 \pm 0.10$ \\
{\scshape Dirichlet FM}   & $2.25 \pm 0.01$  & $2.25 \pm 0.02$ \\
\sflow (ours)    & $\mathbf{1.4\pm0.1}$   & $\mathbf{1.4\pm0.66}$ \\ 
\bottomrule
\end{tabular}
}
\label{tab:enhancer}
    }
\end{table}
\cut{\begin{wraptable}{r}{0.5\textwidth}
    \centering
    \vspace{-10em}
     \caption{\small MSE of the transcription profile conditioned on generated promoter DNA sequences over the test set. The last $3$ MSE and PPL values are from $5$ independent experiments. Other numbers are from~\citet{stark2024dirichlet}.}
    \resizebox{0.5\textwidth}{!}{
    \begin{tabular}{@{}lcc@{}} 
        \toprule
        \textbf{Model} & \textbf{MSE} ($\downarrow$) &  \textbf{PPL} ($\downarrow$) \\
        \midrule
        {\scshape Bit Diffusion (bit-encoding)} & $0.041$ & --- \\
        {\scshape Bit Diffusion (one-hot encoding)} & $0.040$ & --- \\
        {\scshape D3PM-uniform} & $0.038$ & ---\\
        {\scshape DDSM} & $0.033$ & ---\\
        {\scshape Language Model} & $0.034 \pm 0.001$ & $2.247 \pm 0.102$\\
        {\scshape Dirichlet FM} & $0.034 \pm 0.004$ & $\mathbf{1.978 \pm 0.006}$\\ 
        \midrule
        \sflow (ours) & $\mathbf{0.029\pm 0.001}$ & $\mathbf{1.4\pm2.7}$\\ 
        \bottomrule
    \end{tabular}
    }
    \label{tab:promoter}
\end{wraptable}
}
We assess the ability of \sflow to generate DNA sequences. Promoters are DNA sequences that determine where on a gene DNA is transcribed into RNA; they contribute to determining how much transcription happens~\citep{haberle2018eukaryotic}.  The goal of this task is to generate promoter DNA sequences conditioned on a desired transcription signal profile. Solving this problem would enable one to control the expression level of any synthetic gene, \eg in the production of antibodies. For a detailed dataset background, see \S F.1 in~\citet{avdeyev2023dirichlet}.

\looseness=-1
\xhdr{Results} Our experimental evaluation closely follows prior work~\citep{avdeyev2023dirichlet, stark2024dirichlet}. We report the MSE between the signal of our conditionally generated sequence and the target one, a human genome promoter sequence (MSE in~\Cref{tab:promoter}), both given by the same pre-trained Sei model~\citep{chen2022sequence}. We train our model on $88{,}470$ promoter sequences, each of length $1{,}024$, from a database of human promoters~\citep{hon2017atlas}, each sequence having an associated expression level indicating the likelihood of transcription at each DNA position. As shown in \Cref{tab:promoter}, \sflow outperforms baseline methods DDSM~\citep{avdeyev2023dirichlet} and {\scshape Dirichlet FM}~\citep{stark2024dirichlet} on the MSE evaluation. Perplexities (PPL) from \sflow on the test set are also better than the baselines and, on average, improve over {\scshape Dirichlet FM}. 

\cut{
The test set comes from chromosomes 8 and 9 and chromosome 10 is used for the validation set. The test set is comprised of $40,000$. promoter sequences. The flow model used is a $1$-d CNN. See~\S\ref{app:promoter_exp} for further details on the experimental setup.
}

\subsection{Enhancer DNA design}
\cut{
\begin{wraptable}{r}{0.5\textwidth}
\centering
\vspace{-15pt}
\caption{\small perplexities (PPL) values for different methods for enhancer DNA generation. Lower PPL is better. Values are an average and standard error over $5$ seeds.}
\vspace{-5pt}
\resizebox{0.5\textwidth}{!}{%
\begin{tabular}{@{}lcc@{}}
\toprule
\textbf{Method}           & \textbf{Melanoma PPL} ($\downarrow$)   & \textbf{Fly Brain PPL} ($\downarrow$) \\ 
\midrule
Random Sequence   & $895.88$ & $895.88$ \\
Language Model    & $2.22 \pm 0.09$  & $2.19 \pm 0.10$ \\
{\scshape Dirichlet FM}   & $2.25 \pm 0.01$  & $2.25 \pm 0.02$ \\
\sflow (ours)    & $\mathbf{1.4\pm0.1}$   & $\mathbf{1.4\pm0.66}$ \\ 
\bottomrule
\end{tabular}
}
\vspace{-10pt}
\label{tab:enhancer}
\end{wraptable}
}
\looseness=-1
Enhancers are DNA sequences that regulate the transcription of DNA in specific cell types (\eg melanoma cells). Prior work has made use of generative models for designing enhancer DNA sequences in specific cells~\citep{taskiran2024cell}. Following~\citet{stark2024dirichlet}, we report the perplexity over sequences as the main measure of performance. We also include results on Fr\'echet Biological Distance (FBD) with pre-trained classifiers provided in {\scshape Dirichlet FM}~\citep{stark2024dirichlet}, cf.\  \S\ref{app:enhancer_exp}. Nevertheless, those classifiers perform poorly on cell-type classification of Enhancer sequences, with test set accuracies of $11.5\%$ and $11.2\%$ on the Melanoma and FlyBrain datasets, respectively; thus, metrics derived from these are not representative of model quality, which we still included in~\S\ref{app:enhancer_exp} for transparency.



\looseness=-1
\xhdr{Results} 
We report our results in \Cref{tab:enhancer}, with FBD reported in \Cref{tab:enhancer_app}. We observe that \sflow obtains significantly better performance than {\scshape Dirichlet FM}, which highlights its ability to fit the distribution of Melanoma and FlyBrain DNA enhancer sequences. Moreover, we also note that our method improves over the language model baseline on both datasets, which bolsters the belief that \sflow can be used in similar settings to those of autoregressive models.

\subsection{\emph{De novo} molecule generation}

\begin{figure}
\centering
\begin{minipage}[t]{0.45\textwidth}
    \captionof{table}{\small Results on QM9. Higher is better. The baseline numbers are taken from the cited papers. The numbers reported for FlowMol are those for the uniform distribution and end-point parameterisation. Our numbers are for $1{,}000$ molecules.}
    \label{tab:qm9}
    \resizebox{1\textwidth}{!}{
    \begin{tabular}{llll}
    \toprule
     \textbf{Method} & \textbf{Atoms S} (\%) & \textbf{Mols Val} (\%) & \textbf{Mols. S} (\%) \\
     \midrule
     {\scshape Fisher-Flow} (ours) & $98.6$ & $95.3$ & $88.2$ \\
     JODO \citep{jodo} & $99.4$ & $98.9$ & $98.7$ \\
     EquiFM \citep{equifm} & $99.4$ & $94.4$ & $93.2$ \\
     FlowMol \citep{dunn2024mixed} & $98.9$ & $96.9$ & $84.2$ \\
     \bottomrule
    \end{tabular}
    }
\end{minipage}
\hfill
\begin{minipage}[t]{0.45\textwidth}
    \centering
    \caption{{\small Generated molecules using \sflow on QM9.}}
    \includegraphics[width=0.5\linewidth]{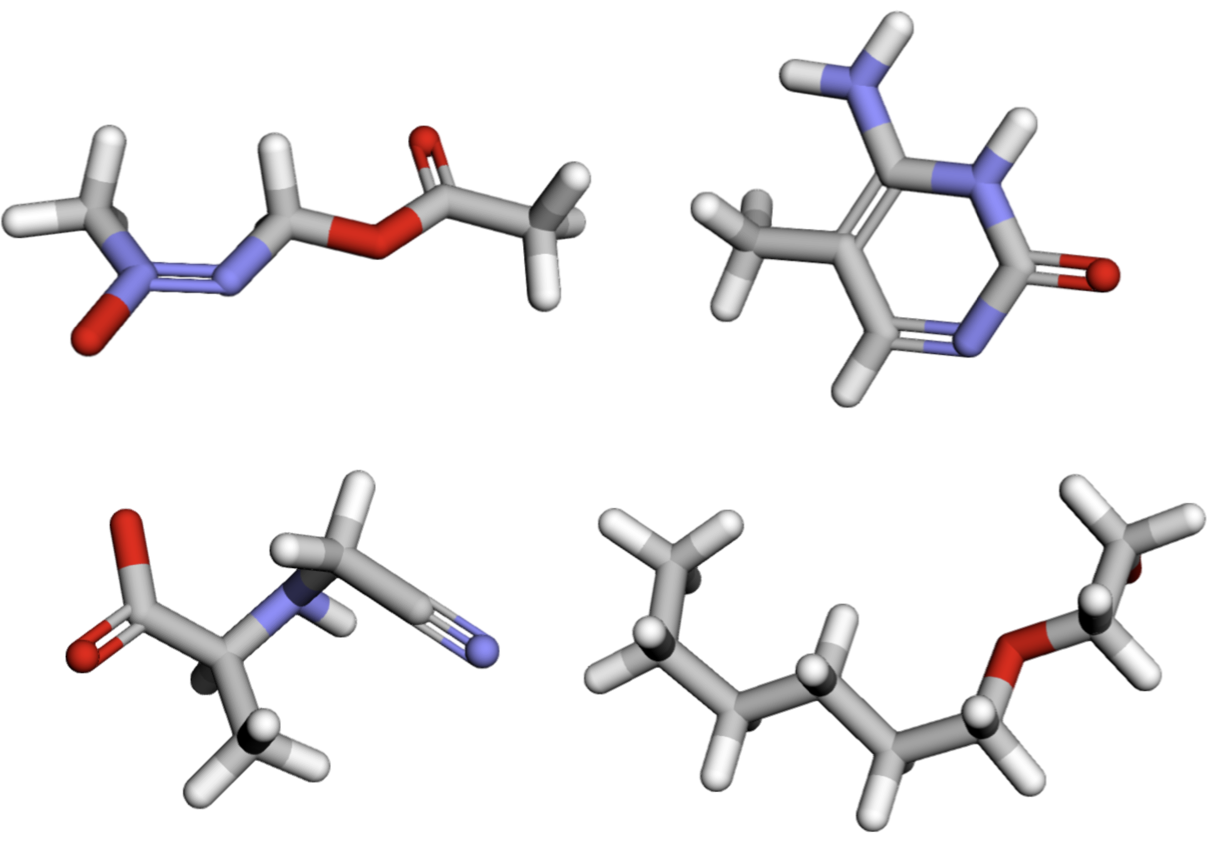}
    \label{fig:samples_qm9}
\end{minipage}
\vspace{-1em}
\end{figure}

In this experiment, we evaluate \sflow's ability to generate molecules unconditionally, a.k.a. ``\emph{de novo}''. The difficulty in this task is that we are interested in generating the positions of the molecules, their atom types, their charges, and the bonds between these, resulting in a high dimensional space with both discrete and continuous data $(\R^d)^n \times (\Delta^a)^n \times (\Delta^c)^n \times (\Delta^e)^{n^2}$, where $n \in \mathbb{N}^\star$ is the number of atoms, $a$ possible atom types, $c$ charges, and $e$ bonds. We train our model over the QM9 dataset \citep{ruddigkeit2012qm9,ramakrishnan2014qm9}. We report the percentage of stable atoms within molecules, valid molecules, and stable molecules. Our implementation is mostly based on that of \citep{dunn2024mixed}.

\begin{figure}
\end{figure}
\xhdr{Results} We report our results in \Cref{tab:qm9}. We also provide some qualitative examples in \Cref{fig:samples_qm9}. As we can see, \sflow compares well on all metrics to {\scshape Simplex-Flow} on all metrics. Nonetheless, it must be reported that the latter, trained with a Gaussian prior, endpoint parameterisation and cosine time schedule performed substantially better than both flow-based methods, closing the gap with the other baselines. It is likely that a more extensive exploration of priors, time parameterisations and other hyperparameters would increase \sflow's performance.

\subsection{Language modelling}
\begin{wraptable}{r}{0.5\textwidth}
    \vspace{-1.25em}
    \centering
        \caption{{\looseness=-1 Test perplexities on the LM1B dataset. All baselines are taken from concurrent work MDLM by \citet{sahoo2024simple}. Best diffusion or flow-matching method is in bold font.}}
    \resizebox{0.5\textwidth}{!}{
    
    \begin{tabular}{llll}
        \toprule
        & \textbf{Method} & \textbf{Parameters} & \textbf{PPL} ($\downarrow$) \\
        \midrule
  \multirow{5}{*}{\rotatebox[origin=c]{90}{Diffusion}} &  {\scshape BERT-Mouth} & 110M  & $\leq 142.89$\\
        &  {\scshape D3PM (absorb)} & 70M  & $\leq 77.50$\\
        & {\scshape Diffusion-LM} & 80M  & $\leq 118.62$\\
        & {\scshape DiffusionBert} & 110M  & $\leq 63.78$\\
        &{\scshape SEDD (33B tokens)} & 110M  & $\leq 32.79$\\
        \midrule
       \multirow{2}{*}{\rotatebox[origin=c]{90}{AR}} & {\scshape Transformer (33B tokens)} & 110M  & $22.32$\\
       & {\scshape Transformer (327B tokens)} & 110M  & $20.86$\\
        \midrule
          \multirow{4}{*}{\rotatebox[origin=c]{90}{DM/FM}}  &{\scshape MDLM (33B tokens)}  & 110M  & $\leq 27.04$\\
        & {\scshape Fisher-Flow (33B tokens)} (ours) & 110M  & $\mathbf{\leq 26.51}$\\
        &{\scshape MDLM (327B tokens)}  & 110M  & $\leq 23.00$\\
        & {\scshape Fisher-Flow (327B tokens)} (ours) & 110M  & $\mathbf{\leq 22.42}$\\
        \bottomrule
    \end{tabular}
    }
    \label{tab:language}
\end{wraptable}
\looseness=-1
Finally, we test the language modelling capabilities of \sflow. To do so, we train the model on the LM1B dataset \citep{chelba2014lm1b}, a large language modelling dataset containing about $800{,}000$ words. For this experiment, we extend \sflow to a masked path as is done by \citep{sahoo2024simple,shi2024simplified}: we define the probability path as $p_t = \kappa_t p_M + (1 - \kappa_t)p_{\mathrm{unif}} $, where $\kappa:[0,1]\to[0,1]$ is a noise scheduler. Here, $p_M$ is the Fisher-Rao geodesic between the target, $x_0$, and the designated mask token $M$, while $p_{\mathrm{unif}}$ is also a Fisher-Rao geodesic between a sample from a uniform distribution and $x_0$. It is thus a convex combination of probability paths. Using a denoising architecture enables us to rewrite the original loss as a weighted negative log likelihood $-\mathbb{E}[\log p(x_0\mid x_t)]$. This allows us to calculate an upper bound on the test perplexity, a natural evaluation metric for language modelling \citep{sahoo2024simple,shi2024simplified}.

\xhdr{Results} The results are given in \Cref{tab:language}. As one can observe, using the Fisher-Rao metric enables better performance than MDLM. Yet, the gap with auto-regressive methods is still significant.

\section{Related work}
\label{sec:rw}
\vspace{-4pt}

\xhdr{Geometric generative models}
There are several methods for defining generative models over Riemannian manifolds, the most pertinent to this work include diffusion models~\citep{huang2022riemannian,de2022riemannian}, normalising flows~\citep{bose2020latent,mathieu2020riemannian,ben2022matching,chen2023riemannian}.
For molecular tasks that require generating nodes and edges, equivariant variants of diffusion and flow-based models are a natural choice~\citep{hoogeboom2022equivariant,xu2022geodiff}.

\looseness=-1
\xhdr{Discrete diffusion and flow models}
Discrete generative models diffusion and flow models can be categorised into either relaxations to continuous spaces~\citep{li2022diffusion,chen2022analog}, or methods that use continuous-time Markov chains with sophisticated transition kernels~\citep{austin2021structured,zhao2024improving,campbell2024generative,lou2023discrete}, with some matching autoregressive models~\citep{gulrajani2024likelihood}. Defining discrete data on the simplex has also been explored in the context of generative models~\citep{han2022ssd,mahabadi2023tess,stark2024dirichlet}. \sflow is fundamentally different from existing works~\citep{austin2021structured,campbell2024generative,alamdari2023protein,lou2023discrete} in that we consider a continuous relaxation of the discrete space and construct vector fields on $\posOrthant^d$. Finally, concurrent to our work~\citet{dunn2024mixed} propose simplex flow matching, and \citet{boll2024generative} introduced $e$-geodesic flows that leverage the Fisher-Rao metric on the assignment manifold~\citep{boll2024generative}. Simplex flow-matching differs from \sflow in that it does not make use of the Fisher-Rao metric. We include a detailed comparison between \sflow in relation to DFM and $e$-Geodesic Flow Matching~\citep{boll2024generative} in~\S\ref{app:dirichlet_fm}.

\section{Conclusion}
\label{sec:conclusion}
\vspace{-4pt}

\looseness=-1
In this paper, we introduce \sflow a novel generative model for discrete data. Our approach offers a novel perspective and reparameterises discrete data to live on the positive orthant of a $d$-hypersphere, which allows us to learn categorical densities by performing Riemannian flow matching. Empirically, \sflow improves performance on synthetic and biological sequence design tasks over comparative discrete diffusion and flow matching models while being more general as a framework. While \sflow enjoys favorable theoretical properties with strong empirical performance, our method is not fully developed for language modeling domains. Consequently, a natural direction for future work is to design variations of \sflow capable of handling larger sequence lengths and discrete categories as found in language domains.

\section*{Acknowledgements}
We thank Alexander Tong for his generous time, help, and guidance in helping with the language modeling experiments. OD is supported by both Project CETI and Intel. MP is supported by CenIA and by Chilean Fondecyt grant n. 1210426. AJB is partially supported by an NSERC Post-doc fellowship. This research is partially supported by EPSRC Turing AI World-Leading Research Fellowship No. EP/X040062/1 and EPSRC AI Hub on Mathematical Foundations of Intelligence: An "Erlangen Programme" for AI No. EP/Y028872/1

\clearpage
\bibliographystyle{abbrvnat}
\bibliography{bibliography}

\clearpage
\appendix
\section{Broader Impacts}
\label{sec:broader_impact}
We would like to emphasise that our paper is mainly theoretical and establishes generative modeling of discrete data using flow matching by continuously reparameterising points onto a statistical manifold equipped with the Fisher-Rao metric. However, more broadly discrete generative modeling based on diffusion models and flow matching has important implications in various fields. In biology, these models enable the generation of novel biological sequences, facilitating the development of new therapeutics. However, the same technology poses risks if exploited maliciously, as it could be used to design harmful substances or biological weapons. In language modeling, the capability to generate coherent and contextually relevant text can significantly enhance productivity, creativity, and communication. Nevertheless, the advent of superhuman intelligence through advanced language models raises concerns about potential misuse, loss of human control, and ethical dilemmas, highlighting the need for robust oversight and ethical guidelines.
\section{Geometry of the Simplex}
\label{app:geometry_simplex}
We introduce here very briefly properties of geometry on the simplex that we use in this paper. Our main reference for these results is \citet{astrom2017simplex}. Note that our implementation for most of these properties relies on that of \citet{JuliaManifolds2023}, which we port to Python. Recall that a $d$-simplex, for $d \in \Nstar$, is defined as $\Delta^d \coloneqq \{x \in \R^{d+1} | \mathbf{1}^\top x = 1, x \geq 0\}$. When equipped with the Fisher-Rao metric, it becomes a Riemannian manifold that is isometric to the positive orthant of the $d$-sphere of in $\R^{d+1}$. That is to say, $\psi:\Delta^d\to \posOrthant^d, (x_0,\dots,x_d)\mapsto (2\sqrt{x_0},\dots,2\sqrt{x_d})$ is a diffeomorphism, where $\posOrthant^d \coloneq \{x \in \R^{d+1} : \|x\|_2=2, x \geq 0\}$; we call $\psi$ the ``sphere-map''.

In the following, $\mathring{\Delta}^d$ denotes the interior of the simplex, and $\tangent_p \Delta^d \coloneqq \{x \in \R^{d+1} : \mathbf{1}^\top x = 0\}$ the tangent space at point $p$. The exp map on the simplex is given by, for all $p \in \mathring{\Delta}^d$, $v \in \tangent_p \Delta^d$,
\begin{equation}
    \exp_{p}(v) = \frac 12 \left(p + \frac{v_p^2}{\lVert v_p^2\rVert^2}\right) + \frac 12 \left(p - \frac{v_p^2}{\lVert v_p^2\rVert^2}\right)\cos(\lVert v_p \rVert) + \frac{\sqrt{p}}{\lVert v_p \rVert}\sin(\lVert v_p \rVert),
\end{equation}
where $v_p \coloneqq \frac{v}{\sqrt{p}}$, and squares, square roots and quotients of vectors are meant element-wise. Similarly, the log map is given by, for $p, q \in \mathring{\Delta}^d$,
\begin{equation}
    \log_{x_0}(x_1) = \frac{d_{\Delta^d}(p, q)}{\sqrt{1 - \langle\sqrt{p}, \sqrt{q}\rangle}}\left(\sqrt{pq} - \langle\sqrt{p},\sqrt{q}\rangle p\right),
\end{equation}
where the product is meant element-wise, and the distance is
\begin{equation}
    d_{\Delta^d} = 2\arccos(\langle\sqrt{p}, \sqrt{q}\rangle).
\end{equation}
The Riemannian metric at point $p \in \mathring{\Delta}^d$ for vectors $u, v \in \tangent \Delta^d$ is given by
\begin{equation}
    \langle u, v\rangle_p = \left\langle\frac{u}{\sqrt{p}},\frac{v}{\sqrt{p}}\right\rangle.
\end{equation}
Finally, for parallel transport, we use the sphere-map, perform parallel-transport on the sphere, and invert the sphere-map.

The relevance of the Fisher-Rao metric stems from the following two characterisations:
\begin{itemize}[noitemsep,topsep=0pt,parsep=0pt,partopsep=0pt,label={\large\textbullet},leftmargin=*]
    \item \emph{The Fisher-Rao metric is the leading-order approximation of the Kullback-Leibler divergence \citep{amari2016information, ay2017information}.} Recall the general setting: if a $d$-dimensional manifold of probability densities $\mathcal M^d$ is parameterised by a differentiable map $\theta\mapsto p_\theta$ from a submanifold $\Theta\subseteq \mathbb R^D$ (note that the requirement $D=d$ is not necessary for the following computations to make sense), then for fixed $\theta_0\in \Theta$ we may Taylor-expand 
    \[
        p(\theta)=p(\theta_0) + \sum_{j=1}^D(\theta^j - \theta_0^j)\frac{\partial p(\theta_0)}{\partial\theta^j} + o(|\theta-\theta_0|),
    \]
    and a straightforward computation gives
    \begin{eqnarray*}
        D_{KL}(p(\theta_0)||p(\theta)) &=&\frac12 \sum_{j,k=1}^D(\theta^j-\theta_0^j)(\theta^k-\theta_0^k)\left.\mathbb E_{p(\theta)}\left[\frac{\partial\log p}{\partial \theta^j}\frac{\partial\log p}{\partial\theta^k}\right]\right|_{\theta=\theta_0} + o(|\theta-\theta_0|^2)\\
        &:=&\frac12 \sum_{j,k=1}^D g_{jk}(\theta_0)[\theta^j-\theta_0^j,\theta^k-\theta_0^k] + o(|\theta-\theta_0|^2).
    \end{eqnarray*}
    Thus the matrix $g(\theta_0)=(g_{ij}(\theta_0))_{i,j=1}^D$ defines the quadratic form on the tangent space $\gT_{\theta_0}\Theta$ which best approximates $D_{KL}(p(\theta_0)||p(\theta))$ in the limit $\theta\to\theta_0$. In the coordinates $\Theta=\Delta^d\subset \mathbb R^{d+1}$, when we parameterise probabilities over $K=d+1$ classes numbered $0,\dots,d$ via the "tautological" parameterisation $\theta=p$ for $p\in \Delta^d$, (explicitly, in this parameterisation class $i$ has probability $p_\theta(i)=\theta^i=p^i$), then we obtain $\frac{\partial \log p_\theta}{\partial \theta^j}=\frac{1}{p^j}\delta(i=j)$ and
    \[
        g_{jk}(p)=\mathbb E_{p(\theta)}\left[\frac{\partial\log p}{\partial \theta^j}\frac{\partial\log p}{\partial\theta^k}\right]=\sum_{i=1}^{d+1} p^i\frac{1}{p^j}\delta(i=j)\frac{1}{p^k}\delta(i=k)=\frac{1}{p^j} \delta(j=k).
    \]
    Thus $g(p)[u,v]=g_{\rm{FR}}(p)[u,v]=\sum_{i=1}^{d+1}\frac{u^iv^i}{p^i}$ as before.
    \item \emph{The Fisher-Rao metric is up to rescaling, the only metric that is preserved under sufficient statistics.} First, for $2\leq K'\leq K$, a define a map $M:\mathcal P([K'])\to\mathcal P([K])$ to be a \emph{Markov map} if there exist probability measures $q_1,\dots, q_{K'}\in \mathcal P([K])$ such that for $p\in \mathcal P([K'])$ we have $M(p)=\sum_{k=1}^{K'}p(k) q_k$. In other words, representing probability spaces as simplices and denoting $d=K-1, d'=K'-1$, we have that $M$ is a Markov map if the simplex $\Delta^{d'}$ is affinely mapped under $M$ to a $d'$-dimensional simplex in $\Delta^d$ (the vertices of the image simplex have been denoted above by $q_1,\dots, q_{K'}$). 
    
    Then a restatement of Chentsov's theorem \cite[Thm. 11.1]{cencov2000statistical}, \cite[Thm. 1.2]{ay2017information} is that if a sequence of Riemannian metrics $g_d$ over $\Delta^d$ defined for $d\geq 2$ satisfies the property that for any $1\leq d'\leq d$ any Markov morphism $M:\Delta^{d'}\to \Delta^d$ is an isometry with respect to metrics $g_{d'},g_d$, then there exists $C>0$ such that each of the $g_d$ is $C$ times the Fisher-Rao metric on $\Delta^d$.

    A common reformulation, interpreting the Markov map reparameterisations $M:\mathcal P([K'])\to\mathcal P([K])$ of $\mathcal P([K'])$ as sufficient statistics, is to say that Fisher-Rao metrics are (up to a common rescaling for all $d$) the only metrics that are invariant under mapping probability measures to sufficient statistics. 
\end{itemize}

\section{Details and proofs for \Cref{sec:fr_metric_theory}}
\label{app:frjustif}

We here recall the setup: we are considering a loss function $\mathcal L:\gP(\gM^d)\to \R$, in which $\gM^d$ is a Riemannian manifold, specifically it will be the simplex $\Delta^d$ endowed with a Riemannian metric $g$. Points $p_\omega\in\gM^d$ represent categorical distributions, as $\gM^d$ was obtained from $\gP(\gA)$ by parametrising it with the simplex $\Delta^d$, thus inducing a differentiable structure. 

The space $\gP(\gM^d)$ is then endowed with the Wasserstein distance $W_{2,g}$ induced by the Riemannian geodesic distance of $(\gM^d, g)$. Then $\gP=(\gP(\gM^d), W_{2,g})$ can be given a Riemannian metric structure too, defined as follows \citep{ambrosio2005gradient, villani2009optimal}. For $p\in \gP$ the tangent space $\gT_p\gP$ is identified with the $L^2(p;g)$-closure of the space of vector fields $v:\gM^d\to \gT\gM^d$ which are the gradient of a $C^1_c$-function $\psi:\gM^d \to \R$. Here we have for $v=\nabla \psi\in \gT_p\gP$
\[
    \|v\|_{L^2(p;g)}^2:=\int_{\gM^d}\|v(p_\omega)\|_g^2 dp(p_\omega),
\]
and the corresponding Riemannian tensor induced by $g$ over $v,w\in \gT_p\gP$ is given by 
\[
    g^{\gP}(v,w):=\int_{\gM^d}\langle v(p_\omega),w(p_\omega)\rangle_{g} dp(p_\omega).
\]
For further details see \citet[Ch. 13]{villani2009optimal}.

In order to find a well-behaved metric over $\gM^d$, we start by considering $\gM^d$ (which in our case is the statistical manifold parameterising the space of categorical probabilities $\gP(\gA)$) with the KL-divergence as a comparison tool for its elements. We will use this divergence in order to regularise the gradient descent of a loss function $\mathcal L:\gP(\gM^d)\to\R$, and to do so we introduce the KL-optimum coupling which for $\mu,\nu\in\gP(\gM^d)$ takes the value
\[
    W_{\rm{KL}}y(\mu,\nu):=\min\left\{\mathbb E_{(p_\omega, p_{\omega'})\sim \pi}\left[\mathbb D_{\rm{KL}}(p_\omega||p_{\omega'})\right]:\ \pi\in \gP(\gM^d\times\gM^d) \text{ has marginals }\mu,\nu\right\}.
\]
In words, $W_{\rm{KL}}$ determines the smallest average displacement required for moving $\mu$ to $\nu$, in which displacements between elements of $\gM^d\simeq \gP(\gA)$ are quantified by $\mathbb D_{\rm{KL}}$-divergence.

We then use this distance to regularise the gradient descent of $\mathcal L$, and show that then the gradient descent converges to the Wasserstein gradient flow on $\mathcal L$, for precisely the Wasserstein distance $W_{2,g_{\rm{FR}}}$ induced by the $g_{\rm{FR}}$-metric over $\gM^d$.

Here we consider a Riemannian metric structure $g$ on $\gM^d$, which we assume to be bounded on the interior $\mathring{\Delta}^d$, \ie to have bounded coefficients when expressed in the parametrisation, which is only used in order to give a rough Lipschitz hypothesis on the underlying parametrisations.

\begin{mdframed}[style=MyFrame2]
\begin{restatable}[extended version of \Cref{prop:fisher_kl}]{proposition}{fisherklapp}
\label{prop:fisher_kl_app}
    Assume that $g$ is a bounded Riemannian metric over $\Delta^d$ such that the parametrisation map $\theta\mapsto p=p(\theta):\Theta\to(\gP(\gM^d), W_{2,g})$ is Lipschitz and differentiable
    . Then the "natural gradient" descent of the form:
    \begin{equation}\label{eq:graddesc_app}
    p(\theta_{n+1})\in \argmin\left\{\mathcal L(p(\theta_{n+1})):\ W_{\rm KL}(p(\theta_{n+1}), p(\theta_n))\leq\epsilon\right\}
    \end{equation}
    approximates, as $\epsilon\to 0^+$, the gradient flow of $\mathcal L$ on manifold $(\gP(\gM^d), W_{g_{\rm{FR}},2})$ with metric $g_{\rm{FR}}^\gP$ induced by Fisher-Rao metric $g_{\rm{FR}}$:
    \begin{equation}\label{eq:gradflow_app}
        \frac{\diff}{\diff s}p(\theta(s)) = \nabla_{g_{\rm{FR}}^{\gP}}\mathcal L(p(\theta(s))).
    \end{equation}
\end{restatable}
\end{mdframed}

\begin{proof}
    We restrict the discussion to the case that $p(\theta)$ is supported in the region $\Delta^d_c:=\{x\in \R^d:\ \mathbb 1\cdot x=1,\ x_i\geq c, 1\le i\le d\}$, and the general result can be recovered by taking $c\to 0^+$. Restricted to this set, it is easy to verify that $\mathbb D_{\mathrm{KL}}$ is bounded.
    
    \textbf{Step 1.} Note that by a small modification of the proof, we can apply \citet[Thm. 10.42]{villani2009optimal} to $\Delta^d$ with cost equal to $\mathbb D_{\rm{KL}}$, and obtain that the $W_{\rm{KL}}$-distance between an admissible competitor $p(\theta+\delta\theta)$ in \eqref{eq:graddesc_app} and $ p(\theta_n)$ is realised by a transport plan $T^{\delta\theta}$, such that we have $p(\theta+\delta\theta)=T^{\delta\theta}_\# p(\theta)$. By definition of $W_{\rm{KL}}$ and due to Chebyshev's inequality, for all $C>0$, the set of points $S_C$ that $T^{\delta\theta}$ moves by more than $C\epsilon$ in $\mathbb D_{\rm{KL}}$-distance has $p(\theta)$-measure not larger than $1/C$. Furthermore, $T^{\delta\theta}$ is uniformly bounded over $\Delta^d_c\setminus S_C$ by our initial hypothesis. By approximating this transport plan by a flow (one can adapt the ideas from \eg \citet[Thm. 4.4]{santambrogio2015optimal} for this contstruction) over $S_C$, we can find a vector field $v^{\delta\theta}$ such that $v^{\delta\theta}(p_\omega) = \frac1{\epsilon}\log_{p_\omega}(T^{\delta\theta}(p_\omega)) + o_{\epsilon}(|\delta\theta|)$ for $p_\omega\in S_C$, with error uniformly bounded in $p_\omega\in\gM$. We then extend $v^{\delta\theta}$ arbitrarily outside $S_C$. This procedure associates to each small enough change $\delta\theta$ a vector field $v_{\delta\theta}\in T_{p(\theta)}\gP$ which whose time-$\epsilon$ flow, denoted $\phi_{v_{\delta\theta}}(t=\epsilon,\cdot)$ pushes measure $p(\theta)$ to a measure approaching $p(\theta+\delta\theta)$ in the limit $\epsilon\to 0, C\to\infty$.

    \textbf{Step 2.} We approximate the optimisation problem \eqref{eq:graddesc_app}. For the constraint, we recall that as noted in \Cref{app:geometry_simplex}, we have Taylor expansion $\mathbb D_{\rm{KL}}(p_\omega||p_{\omega'})= \frac12\|\omega-\omega'\|_{g_{\rm{FR}}}^2 + O(\|\omega-\omega'\|^3)$. For approximating $\mathcal L$ we use its differentiability and get $\mathcal L(p(\theta')) = \mathcal L(p(\theta))+d\mathcal L(p(\theta))[v]$, for $v\in \gT_p\gP$. Thus minimisation problem \eqref{eq:graddesc_app} is well approximated, (in the limits mentioned in the previous step) by 
    \begin{equation}\label{eq:gradesc1_app}
        p(\theta_{n+1})=\left(\phi_{v_{\delta\theta}}(1,\cdot)\right)_\# p(\theta_n),\quad v_{\delta\theta}\in\mathrm{argmin}_v\left(\epsilon \ d\mathcal L(p(\theta))[v]:\ \langle v, v\rangle_{g_{\rm{FR}}^{\gP}}=1\right),
    \end{equation}
    in which we used a rescaling compared to previous step, given by $v\mapsto \epsilon v$. This means that we used the associated flow up to time $1$ rather than time $\epsilon$, and thus the minimisation has to be taken amongst elements $v\in T_{p(\theta)}\gP$ and we approximate the constraint by $\langle v, v\rangle_{g_{\rm{FR}}^{\gP}}= 1$, which replaces the correct constraint $W_{\rm{KL}}(p(\theta+\delta\theta), p(\theta))=\epsilon$.
    
    \textbf{Step 3.} In the optimisation \eqref{eq:gradesc1_app}, we have a quadratic constraint over the vector space $T_{p(\theta_n)}\gP$, and thus we can use Lagrange multipliers, and for the optimiser we need to look for critical points of $v\mapsto \epsilon d\mathcal L(p(\theta))[v] +\frac{\lambda}{2}\langle v, v\rangle_{g_{\rm{FR}}^{\gP}}$, in which $\lambda$ is the Lagrange multiplier, to be fixed at the end using the constraint. This gives the following characterisation of the optimiser $v_{\delta\theta}^*$:
    \begin{equation}\label{eq:optimizer}
        \forall w\in T_{p(\theta)}\gP,\quad \langle v_{\delta\theta}^*, w\rangle_{g_{\rm{FR}}^{\gP}} = -\frac{\lambda}{\epsilon}d\mathcal L(p(\theta))[w]\quad\Longleftrightarrow\quad v_{\delta\theta}^*=-\frac{\lambda}{\epsilon}\nabla_{g_{\rm{FR}}^{\gP}}\mathcal L(p(\theta)),
    \end{equation}
    in which we just use the classical definition of the gradient on a manifold.

    This means that in the approximation of $\epsilon\to 0$ the step $p(\theta)\to p(\theta+\delta\theta)$ must move in the negative-$g_{\rm{FR}}^{\gP}$-gradient direction of $\mathcal L$ at $p(\theta)$, as desired.
\end{proof}
\section{Optimal Transport proofs}
\label{app:ot}

\begin{mdframed}[style=MyFrame2]
\begin{restatable}[extended version of \Cref{prop:mongegeodesic}]{proposition}{mongegeodesicapp}
\label{prop:mongegeodesicapp}
    For any two Borel probability measures $p_0, p_1\in\mathcal P(\posOrthant)$, the following hold:
    \begin{enumerate}\item There exists a unique OT-plan $\pi$ between $p_0,p_1$. 
    \item For $t\in[0,1]$ let $e_t(x_0,x_1)$ be the constant-speed parameterisation of the unique geodesic of extremes $x_0$ and $x_1$, defining the map 
    \begin{equation}\label{eq:etapp}
    e_t:\posOrthant\times\posOrthant\to\posOrthant, \quad e_t(x_0,x_1)\coloneqq\exp_{x_0}(t\log_{x_0}(x_1)).
    \end{equation}
    Then there exists a unique Wasserstein geodesic $(p_t)_{t\in[0,1]}$ connecting $p_0$ to $p_1$, and it is given by
    \begin{equation}\label{eq:displconvapp}
        p_t\coloneqq(e_t)_\#\pi\in \gP(\posOrthant),\quad t\in [0,1].
    \end{equation}
    \item For every point $x_t$ in the support of $p_t$, there exists a unique pair $(x_0,x_1)$ in the support of the optimal transport plan $\pi$ such that $x_t=e_t(x_0,x_1)$. Furthermore, the assignment $x_t\mapsto (x_0, x_1)$ is continuous in $x_t$.
    \item The probability path $(p_t)_{t\in[0,1]}$ has velocity field $u_t:=\log_{x_t}(x_1)-\log_{x_t}(x_0)$, which is uniquely determined over the support of $p_t$.
    \item The above probability measure path and associated velocity fields $(p_t, u_t)_{t\in[0,1]}$ are minimisers of the following kinetic energy minimisation problem
    \begin{equation}\label{eq:benamoubrenierapp}
        \min_{(\rho_t, v_t)_{t\in[0,1]}}\left\{\int_0^1\mathbb E_{\rho_t}[\|v_t\|^2] \diff t:\ \partial_t \rho_t + \mathrm{div}(\rho_t v_t)=0, \quad \rho_0=p_0,\ \rho_1=p_1\right\}.
    \end{equation}
    \end{enumerate}
\end{restatable}
\end{mdframed}

\begin{proof}
    For point 1, we can use \citet[Thm. 10.28]{villani2009optimal} (the simpler \citet[Thm. 10.41]{villani2009optimal} also applies, with the minor modification that we work on a manifold with boundary). To verify its conditions, note that $\gM^d\subset\posOrthant$ is a subset of a Riemannian manifold and has $(d-1)$-dimensional measure, and that cost $c(x,y)=d^2(x,y)$ is convex, thus it has unique superdifferential and $\nabla_xc(x,\cdot)$ is injective, as required.

    For points 2 and 3, we note that by \citet[Cor. 7.22]{villani2009optimal} (see also \citet{mccann2001polar}), in general Polish spaces displacement interpolants as given by \eqref{eq:etapp} and \eqref{eq:displconvapp}, coincide with Wasserstein geodesics. 
    
    A simplified version of the proof of 4. is present in \citet[Prop. 5.30]{santambrogio2015optimal}. For the general case, we can use  \citet[Thm. 10.28]{villani2009optimal}, in particular eq. (10.20) therein. Note that for $c(x,y)=d^2(x,y)$, as indicated in Example 10.36 this equation corresponds to the equation of geodesics in the underlying manifold. Then we just note that $u_t$ is the velocity field of a constant speed geodesic.

    Point 5 is a special case of \citet[Thm. 7.21]{villani2009optimal}, see also \citet{granieri2007action}.
     
\end{proof}
\section{Relation to prior work on the simplex}
\label{app:relation_to_prior_work}
\subsection{Dirichlet Flow matching}
\label{app:dirichlet_fm}
In this appendix, we discuss how flow matching can be done on the simplex using Dirichlet conditional probability paths. This recovers the simplex flows designed in ~\citet{stark2024dirichlet, campbell2024generative}.

The equivalent of a uniform density over $\Delta^d$ is given by a Dirichlet distribution with parameter vector $\alpha = \mathbf{1}$, \ie $p_1(x_1) = \text{Dir}(x_1; \alpha = \mathbf{1})$. This is the starting point for defining a flow between our data distribution, $p_0$, and the Dirichlet prior $p_1$. As proven in~\citet{stark2024dirichlet} we can reformulate \eqref{eq:CFM} using a cross-entropy objective, 
\begin{align}
    \gL_{\rm ce}(\theta) & = \mathbb{E}_{t, q(z), p_t(x_t | z)} \|v_\theta(t, x_t) - u_t(x_t | z)\|_g^2 \\
    & =  \mathbb{E}_{t, q(z), p_t(x_t | z)} \|\log \hat{p}_{\theta} (x_0 | x_t)\|_g^2. 
\end{align}
Here, we parameterise a \emph{denoising classifier} which predicts a denoised sample $x_0$ from $x_t$, which is built using the conditioner $z$. Such a parameterisation naturally restricts the vector field to move tangentially to the simplex and also training is simplified as we do not need to explicitly construct the conditional vector field $u_t (x_t | z)$ during training. At inference, we can recover $v_{\theta}(t, x_t) = \sum_i^d u_t(x_t | x_0 = e_i) \hat{p}_{\theta}(x_0 = e_1 | x_t)$ and follow the $v_{\theta}$ by integrating time to $t=1$.

\xhdr{Designing conditional paths}
There are two primary points of attack when designing a flow-matching model. We can either define an interpolant $\psi_t(x_t | z)$ with initial conditions $\psi_0 = x_0$, which we can differentiate to obtain $u_t$, \ie $\dot{\psi}(x_t | z) = u_t(x_t | z)$; or we can operate on the distributional level and specify conditionals $p_t(x_t | z)$ from which a suitable vector field can be recovered.

If we take the interpolant perspective, one can easily implement the linear interpolant ~\citep{lipman_flow_2022,tong2023improving}, which gives the following conditional vector field:

\begin{align}
    & \psi_t(x_t | x_0, x_1) = tx_0 + (1-t) x_1 \\
    & u_t (x_t | x_0, x_1) = \frac{x_t - x_0}{t} = x_0 - x_1
\end{align}

Unfortunately, in the case of flow matching on the simplex, the linear interpolant has undesirable properties in that the intermediate distribution induced by the flow must quickly reduce support over $\Delta^d$ by dropping vertices progressively for $t > 0$~\citep{stark2024dirichlet}. 

Operating directly on the distribution level, we can define $p_t$ as themselves being Dirichlet distributions indexed by $t$ such that, at $t=0$, we have a uniform mass over $\Delta^d$, and that, at $t = 1$, we reach a vertex. One choice of parameterisation that fulfills these desiderata is

\begin{equation}
    p_t (x_t | x_0 = e_i) = \text{Dir}(x_t; \alpha = 1 + t' \cdot e_i),
    \label{eq:dirichlet_conditional}
\end{equation}

where $t' = f(t)$ is a monotonic function of the original time variable $t$ such that $f(0) = 0$ and $\lim_{t \to 1^{-}} f(t) = \infty$. Clearly, $t' =0$ recovers the uniform prior as $\alpha = \mathbf{1}^\top$, while $t' \to \infty$ increases the mass of $e_i$ while other vertices remain constant. Given the conditional in \eqref{eq:dirichlet_conditional}, one corresponding vector field that satisfies the continuity equation is
\begin{equation}
    u_t(x_t | x_0 = e_i) = C(x_i, t)(x_t - e_i), \quad  C([x_t]_i, t) = -\tilde{I}_{x_i}(t+1, d-1)\frac{\gB(t+1, d-1)}{(1-x_i)^{d-1}x^t_i},
\end{equation}
\looseness=-1
where $\tilde{I}_{x}(a,b) = \frac{\partial}{\partial a}I_x(a,b)$ is the derivative of the regularised incomplete beta function~\citep[Appendix A.1]{stark2024dirichlet} and $C \propto 1/t$ as in regular linear flow matching.

\subsection{$e$-geodesics on the Assignment manifold}
\label{app:e_geodesics_comparison}

In this appendix, we survey other common geometries implied by the theory of $\alpha$-divergences on statistical manifolds, described in more detail in  \citet[Ch. 4]{amari2016information} or \citet[Ch. 2]{ay2017information}, of which the case of $e$-connections was proposed in relation to flow-matching in \citet{boll2024generative}. 

In what has become a fundamental paper for the field of Information Geometry, \citet{amari2009alpha} unified several commonly used parameterisations of statistical manifolds, in the theory of so-called $\alpha$-connections. Without entering full details (which can be found in the mentioned references), on a statistical manifold, endowed with Fisher-Rao metric, one can introduce a $1$-parameter family of affine connections, so-called $\alpha$-connections with $\alpha\in[-1,1]$, where $\alpha=0$ corresponds to Fisher-Rao Levi-Civita connection, and other notable values are the $m$-connection for $\alpha=-1$ and the $e$-connection for $\alpha=1$. Furthermore, specific classes of $\alpha$-divergences -- which for $\alpha=0$ recover KL divergence -- have been introduced as adapted to the corresponding $\alpha$-connections.

In general, a choice of differential geometric connection allows to define ad-hoc covariant derivatives, and corresponds to an explicit formula for associated geodesics (curves whose tangent vector has zero covariant derivative). 

For the case of $\alpha$-connections on categorical probabilities $\gP(\gA)$, explicit formulas can be given (see \citet[Ch. 2]{ay2017information}), recovering, for $m$-connections, interpretations as mixtures, with geodesics equal to straight lines in $\Delta^d$-parameterisation, and for $e$-connections geodesics can be interpreted as exponential mixtures, as elucidated in \citet[Ch. 2]{ay2017information} and illustrated in \citet{boll2024generative}.

For the case of $e$-connections, concurrent work \citep{boll2024generative} has proposed to use the corresponding explicit parameterisation of geodesics in flow-matching, leaving as an open question the adaptation of Optimal Transport ideas to the framework.
\section{Implementation Details}
\label{app:implementation_details}
\subsection{General Remarks}
All of our code is implemented in Python, using \verb|PyTorch|. For the implementation of the manifold functions (such as $\log$, $\exp$, geodesic distance, etc.), we have tried two different versions. The first one was a direct port of \verb|Manifolds.JL| \citep{JuliaManifolds2023}, originally written in Julia; the second one used the \verb|geoopt| library \citep{geoopt2020kochurov} as a back-end. The latter performed noticeably better---the underlying reason being probably a better numerical stability of the provided functions.

As for the optimal transport part, it is essentially an adaptation of that of FoldFlow \citep{bose2023se}, which itself relies on the \verb|POT| library \citep{flamary2021pot}.

\subsection{\sflow Algorithm}
\label{app:algorithm_box}
We provide pseudo-code for training \sflow~\Cref{alg:sflow_training}. 

\begin{algorithm}[t]
\caption{\sflow, training on $\posOrthant^d$.}
\label{alg:sflow_training}
\begin{algorithmic}[1]
\State \textbf{Input:} Source and target distributions, $p_1, p_0$, flow network $v_\theta$.
\While{Training}
    \State $t, x_0, x_1 \sim \gU(0, 1), p_0, p_1 = p_{\mathrm{data}}$
    \State $\bar{\pi} \gets \text{OT}_{\posOrthant^d}(x_0, x_1)$ \Comment{Since $x_1$ is one-hot encoded, it is on $\posOrthant^d$.}
    \State $x_0, x_1 \sim \bar{\pi}$
    \State $x_t \gets \exp_{x_0}(t \log_{x_0}(x_1))$\Comment{Geodesic interpolant between $r_0, r_1 \in \posOrthant^d$.}
    \State $u_t(x_t|x_0, x_1) \gets \dot{x}_t$\Comment{Calculated either explicitly or with a numerical approximation.}
    \State $\gL_{\sflow} \gets \norm*{v_\theta(t, x_t) - u_t(x_t|x_0, x_1)}_{\posOrthant^d}^2$
    \State $\theta \gets \text{Update}(\theta, \nabla_\theta \gL_{\sflow})$
\EndWhile
\State \textbf{return} $v_\theta$
\end{algorithmic}

\end{algorithm}

\subsection{Compute Resources}
\label{app:compute_resources}
All experiments are run on a single Nvidia A10 or RTX A6000 GPUs.

\subsection{Experiments}
\subsubsection{Toy Experiment}
\label{app:toy_experiment}
We reproduce most hyper-parameters, except for the number of epochs trained for $500$ instead of $540{,}000$. Nonetheless, a \emph{major} modification from the original setting is the size of the dataset. Indeed, in the original dataset code of \citet{stark2024dirichlet}\footnote{\url{https://github.com/HannesStark/dirichlet-flow-matching/blob/main/utils/dataset.py\#L53}, retrieved on \today.}, one can observe that the points are generated at each retrieval, and the defined length of the dataset is of $10^9$, thus amounting to $540{,}000\cdot 10^9$ training points by the end of the training process. This results in an unrealistic learning setup. To slightly toughen the experiment, we limit the training set size to $100{,}000$ points.

Note that the model with which we train our method is a much simpler architecture than that of {\scshape Dirichlet FM} (which was the one used in \citet{stark2024dirichlet}), ours consisting exclusively of (residual) MLPs. For lower dimensions, it has less parameters, and slightly more in higher dimensions. The other baselines were run with our MLP too.

\begin{table}[ht]
\centering
\vspace{-10pt}
\caption{\small Fr\'echet Biological Distance (FBD) and perplexities (PPL) values for different methods for enhancer DNA generation. Lower FBD and PPL are better. Values are an average and standard error over $5$ different runs.}
\resizebox{\textwidth}{!}{%
\begin{tabular}{@{}lcccc@{}}
\toprule
\textbf{Method}          & \textbf{Melanoma FBD} ($\downarrow$) & \textbf{Melanoma PPL} ($\downarrow$) & \textbf{Fly Brain FBD} ($\downarrow$) & \textbf{Fly Brain PPL} ($\downarrow$) \\ 
\midrule
Random Sequence & $619.0 \pm 0.8$ & $895.88$ & $832.4 \pm 0.3$ & $895.88$ \\
Language Model  & $35.4 \pm 0.5$  & $2.22 \pm 0.09$ & $25.7 \pm 1.0$ & $2.19 \pm 0.10$ \\
{\scshape Dirichlet FM}    & $\mathbf{7.3 \pm 1.2}$ & $2.25 \pm 0.01$ & $6.8 \pm 1.8$ & $2.25 \pm 0.02$ \\
\sflow (ours)   & $27.5\pm2.6$ & $\mathbf{1.4\pm0.1}$ & $\mathbf{3.8 \pm 0.3}$ & $\mathbf{1.4\pm0.66}$ \\ 
\bottomrule
\end{tabular}
}
\vspace{-15pt}
\label{tab:enhancer_app}
\end{table}

\subsubsection{Promoter DNA}
\label{app:promoter_exp}

We train our generative models for $200{,}000$ steps with a batch size of $256$. We cache the best checkpoint over the course of training according to the validation $MSE$ between the true promoter signal and the signal from the Sei model conditioned on the generated promoter DNA sequences. We use the same train/val/test splits as~\citet{stark2024dirichlet} of size $88{,}470$/$3{,}933$/$7{,}497$.

The generative model used for \sflow and DFM~\citet{stark2024dirichlet} is a $20$ layer $1$-d CNN with an initial embedding for the DNA. Each block consists of a LayerNorm~\citep{ba2016layer} followed by a convolutional layer with kernel size $9$ and ReLU activation and a residual connection. As we stack the layers we increase the dilation and padding of the convolutional allowing the receptive field to grow~\citep{oord2016wavenet}. In general, we use the AdamW optimiser~\citep{loshchilov2019adamw}.

Our Language Model implementation is identical to~\citet{stark2024dirichlet} and we use the pre-trained checkpoint provided by the authors and evaluated on the test set.

\subsubsection{Enhancer DNA}
\label{app:enhancer_exp}

We consider two DNA enhancer datasets, the fly brain enhancer dataset with $81$ classes~\citep{janssens2022decoding}, the classes are different cell types, and the melanoma enhancer dataset with $47$ classes~\citep{atak2021interpretation}. Both datasets are comprised of DNA sequences of length $500$. We use the same train/val/test splits as~\citet{stark2024dirichlet} of size $70{,}892$/$8{,}966$/$9{,}012$ for the human melanoma and $83{,}726/10{,}505/10{,}434$ for the fly brain enhancer DNA dataset.

The generative model for our experiments for \sflow and DFM~\citep{stark2024dirichlet} is the same as used for our promoter DNA experiments. Specifically, we use a $20$ layer $1$-d CNN with an initial embedding for the DNA. Each block consists of a LayerNorm~\citep{ba2016layer} followed by a convolutional layer with kernel size $9$ and ReLU activation followed by a residual connection. As we stack the layers we increase the dilation and padding of the convolutional allowing the receptive field to grow~\citep{oord2016wavenet}.

We train for a total of $450{,}000$ steps with a batch size of $256$ where we cache the best checkpoint according to the validation FBD. The test set results in~\Cref{tab:enhancer} are using the best checkpoint according to the validation FBD.

To calculate the FBD we compare embeddings from \sflow with a shallow $5$ layer classifier embeddings originally trained to classify the cell type given the enhancer DNA sequences. Our Language Model implementation is identical to~\citet{stark2024dirichlet} and we use the pre-trained checkpoint provided by the authors and evaluated on the test set.

\subsection{Additional Metrics}
\label{app:additional_metrics}

For complete transparency, we also report the Fr\'echet Biological Distance (FBD) for the DNA enhancer generation experiments as initially reported in~\citet{stark2024dirichlet}. 

The FBD computes Wasserstein distance between Gaussians of embeddings generated and training samples under a pre-trained classifier trained to predict cell-type of enhancer sequences. Versus those embeddings from the generative model under consideration. So crucially there is a dependence on classifier features. 

On FlyBrain we find that \sflow also improves over DFM in FBD being roughly $\approx 2 \times$ better while DFM is better on Melanoma. However, we caveat both FBD results by noting the trained classifiers provided in DFM~\citep{stark2024dirichlet} obtain a test set accuracy of $11.5\%$ and $11.2\%$ on the Melanoma dataset and FlyBrain dataset respectively. Moreover, switching out the pre-trained classifier for another trained from scratch caused large variations in FBD metrics. As a result, the low-accuracy classifiers do not provide reliable representation spaces needed to compute FBD metrics. Consequently, FBD in this setting is a noisy metric that is loosely correlated with model performance, so we opt to report perplexities in~\Cref{tab:enhancer}.

\subsection{\textit{De Novo} Molecule Generation}
Following the setup of \citet{dunn2024mixed}, we report the following metrics for our model on \textit{de novo} molecule generation over the QM9 and GeomDrugs datasets: percentage of stable atoms, percentage of stable molecules, percentage of valid molecules. Note that the following inference scheme is used, when training a model $\hat{x}_1$ on endpoint prediction:
\begin{equation}
x_{t+1} = \exp_{x_t}\left( \alpha'(t) \frac{\Delta}{1-\alpha(t)} \log_{x_t}(\hat{x}_1) \right),
\end{equation}
where $\Delta = 1/N$, and $N > 0$ is the number of integration steps.

\newpage
\section*{NeurIPS Paper Checklist}

\begin{enumerate}

\item {\bf Claims}
    \item[] Question: Do the main claims made in the abstract and introduction accurately reflect the paper's contributions and scope?
    \item[] Answer: \answerYes{} 
    \item[] Justification: The claims made in the abstract and introduction are backed up by our theoretical contributions and experiments results.

\item {\bf Limitations}
    \item[] Question: Does the paper discuss the limitations of the work performed by the authors?
    \item[] Answer: \answerYes{} 
    \item[] Justification: We discuss limitations in the conclusion. In particular that we have not developed our method for language tasks.

\item {\bf Theory Assumptions and Proofs}
    \item[] Question: For each theoretical result, does the paper provide the full set of assumptions and a complete (and correct) proof?
    \item[] Answer: \answerYes{} 
    \item[] Justification: The theory underpinning our method is backed up by theoretical results outlined in the main paper and elaborated on in the appendix. We have stated all assumptions and referenced relevant prior work.

    \item {\bf Experimental Result Reproducibility}
    \item[] Question: Does the paper fully disclose all the information needed to reproduce the main experimental results of the paper to the extent that it affects the main claims and/or conclusions of the paper (regardless of whether the code and data are provided or not)?
    \item[] Answer: \answerYes{} 
    \item[] Justification: We provide all the details of the experimental setup to reproduce our method in our experiments section and in the appendix. We include our code as a .zip file as supplementary material with instructions to reproduce our results. Our code will be made public upon publication.

\item {\bf Open access to data and code}
    \item[] Question: Does the paper provide open access to the data and code, with sufficient instructions to faithfully reproduce the main experimental results, as described in supplemental material?
    \item[] Answer: \answerYes{} 
    \item[] Justification: We include a .zip file with our code base and the necessary commands to reproduce our experiments. All the datasets we use are open-access.

\item {\bf Experimental Setting/Details}
    \item[] Question: Does the paper specify all the training and test details (e.g., data splits, hyperparameters, how they were chosen, type of optimizer, etc.) necessary to understand the results?
    \item[] Answer: \answerYes{} 
    \item[] Justification: We outline our experimental settings in detail in our~\Cref{app:implementation_details}.

\item {\bf Experiment Statistical Significance}
    \item[] Question: Does the paper report error bars suitably and correctly defined or other appropriate information about the statistical significance of the experiments?
    \item[] Answer: \answerYes{} 
    \item[] Justification: We make a significant effort to produce results with means and standard errors over $5$ different runs with different random seeds. For our method and the main baselines we consider. This is in stark contrast to prior work.

\item {\bf Experiments Compute Resources}
    \item[] Question: For each experiment, does the paper provide sufficient information on the computer resources (type of compute workers, memory, time of execution) needed to reproduce the experiments?
    \item[] Answer: \answerYes{} 
    \item[] Justification: We outline the compute resources required in~\Cref{app:compute_resources}.
    
\item {\bf Code Of Ethics}
    \item[] Question: Does the research conducted in the paper conform, in every respect, with the NeurIPS Code of Ethics \url{https://neurips.cc/public/EthicsGuidelines}?
    \item[] Answer: \answerYes{} 
    \item[] Justification: We reviewed the code of ethics and our research is in line with the code.

\item {\bf Broader Impacts}
    \item[] Question: Does the paper discuss both potential positive societal impacts and negative societal impacts of the work performed?
    \item[] Answer: \answerYes{} 
    \item[] Justification: We consider the broader impact of our work in~\Cref{sec:broader_impact}.
    
\item {\bf Safeguards}
    \item[] Question: Does the paper describe safeguards that have been put in place for responsible release of data or models that have a high risk for misuse (e.g., pretrained language models, image generators, or scraped datasets)?
    \item[] Answer: \answerNA{} 
    \item[] Justification: The paper does not put forward models and data which pose such risks.

\item {\bf Licenses for existing assets}
    \item[] Question: Are the creators or original owners of assets (e.g., code, data, models), used in the paper, properly credited and are the license and terms of use explicitly mentioned and properly respected?
    \item[] Answer: \answerYes{} 
    \item[] Justification: All existing assets which are used are open-source and properly cited.

\item {\bf New Assets}
    \item[] Question: Are new assets introduced in the paper well documented and is the documentation provided alongside the assets?
    \item[] Answer: \answerYes{} 
    \item[] Justification: We do not release any new assets. We will release our documented code base upon publication.

\item {\bf Crowdsourcing and Research with Human Subjects}
    \item[] Question: For crowdsourcing experiments and research with human subjects, does the paper include the full text of instructions given to participants and screenshots, if applicable, as well as details about compensation (if any)? 
    \item[] Answer: \answerNA{} 
    \item[] Justification: Our paper does not involve human subjects.

\item {\bf Institutional Review Board (IRB) Approvals or Equivalent for Research with Human Subjects}
    \item[] Question: Does the paper describe potential risks incurred by study participants, whether such risks were disclosed to the subjects, and whether Institutional Review Board (IRB) approvals (or an equivalent approval/review based on the requirements of your country or institution) were obtained?
    \item[] Answer: \answerNA{} 
    \item[] Justification: Our paper does not involve human subjects.

\end{enumerate}

\end{document}